%% file: main.tex
\documentclass{article}

     \PassOptionsToPackage{numbers, compress}{natbib}

     \usepackage[preprint]{neurips_2021}

\usepackage[utf8]{inputenc} 
\usepackage[T1]{fontenc}    
\usepackage{hyperref}       
\usepackage{url}            
\usepackage{booktabs}       
\usepackage{amsfonts}       
\usepackage{nicefrac}       
\usepackage{microtype}      
\usepackage{xcolor}         
\usepackage{amsthm}
\usepackage{amsmath}
\usepackage{amssymb}
\usepackage{etoolbox}		
\usepackage{xspace}			
\usepackage{algorithm}
\usepackage{algpseudocode}
\usepackage{graphics} 		
\usepackage{graphicx}
\usepackage{minitoc}
\usepackage[toc,page,header]{appendix}
\usepackage{accents}

\input{Matiascommands.sty}
\graphicspath{{./figures/}}

\input{./contents/0_0_Authors_info}

\begin{document}

\maketitle
\doparttoc 
\faketableofcontents 


\input{./contents/0_Abstract}


\section{Introduction}
\input{./contents/1_Introduction}

\section{Problem formulation}
\input{./contents/2_Problem_formulation}

\section{Regret lower bound}
\input{./contents/3_Lower_bound}

\section{The Weighted Thompson Sampling (\WTS) algorithm}
\input{./contents/4_WTS}

\section{Application to gain estimation of dynamical systems}
\input{./contents/5_Applications}

\section{Conclusions}
\input{./contents/6_Conclusions}


\newpage
\medskip
\bibliographystyle{IEEEtran}
\bibliography{nips2021}




\parttoc 
\newpage
\normalsize
\appendix
\addcontentsline{toc}{section}{Appendix} 
\part{Appendix} 
\parttoc 

\section{Proof of Theorem~\ref{thm:lower_bound}: regret lower bound}
\input{./contents/Appendix/thm_lower}

\section{Proof of Lemma~\ref{lem:WTS_suff}: sufficient statistics}
\input{./contents/Appendix/lem_suff}

\section{Proof of Lemma~\ref{lem:post_WTS}: posterior mean distribution}
\input{./contents/Appendix/lem_post}

\section{Proof of Theorem~\ref{thm:upper_WTS_unknown}: regret upper bound}
\input{./contents/Appendix/thm_upper}

\section{Approximating $\rho^t$ via Monte Carlo simulations}
\input{./contents/Appendix/rho}

\section{Details of simulation study 1 in Section~\ref{subsec:WTS_sim}}
\input{./contents/Appendix/simulation}

\end{document}

%% file: contents/0_0_Authors_info.tex
\title{
Asymptotically Optimal Bandits \\ under Weighted Information
}

\author{
  Matias I.~M\"uller
    \\
  Division of Decision and Control Systems\\
  KTH Royal Institute of Technology\\
  SE-100 44 Stockholm, Sweden \\
  \texttt{mimr2@kth.se} \\
   \And
   Cristian R.~Rojas \\
  Division of Decision and Control Systems\\
  KTH Royal Institute of Technology\\
  SE-100 44 Stockholm, Sweden \\
  \texttt{crro@kth.se} \\
}

%% file: contents/0_Abstract.tex
\begin{abstract}

We study the problem of regret minimization in a multi-armed bandit setup where the agent is allowed to play multiple arms at each round by spreading the resources usually allocated to only one arm.
At each iteration the agent selects a normalized power profile and receives a Gaussian vector as outcome, where the unknown variance of each sample is inversely proportional to the power allocated to that arm.
The reward corresponds to a linear combination of the power profile and the outcomes, resembling a linear bandit.
By spreading the power, the agent can choose to collect information much faster than in a traditional multi-armed bandit at the price of reducing the accuracy of the samples.
This setup is fundamentally different from that of a linear bandit---the regret is known to scale as $\Theta(\sqrt{T})$ for linear bandits, while in this setup the agent receives a much more detailed feedback, for which we derive a tight $\log(T)$ problem-dependent lower-bound.
We propose a Thompson-Sampling-based strategy, called Weighted Thompson Sampling (\WTS), that designs the power profile as its posterior belief of each arm being the best arm, and show that its upper bound matches the derived logarithmic lower bound.
Finally, we apply this strategy to a problem of control and system identification, where the goal is to estimate the maximum gain (also called $\mathcal{H}_\infty$-norm) of a linear dynamical system based on batches of input-output samples.
%
\end{abstract}

%% file: contents/1_Introduction.tex
\label{sec:1_Introduction}

A stochastic multi-armed bandits (MAB, or simply bandit) is a fundamental instance of reinforcement learning which originated in the early 1900's~\cite{thomps33}, but it only recognized as an experiment design problem in the 1950's~\cite{robbin52}.
MABs capture an essential sequential decision problem subject to an exploration-exploitation trade-off.
In a general MAB  setup, an agent is presented with a finite set of actions from which it can take only one at every round, revealing a random outcome/pay-off/reward whose mean is fixed and unknown to the agent.
The goal of the agent is to sample arms sequentially so as to maximize the expected cumulative reward or, equivalently, to minimize the regret of not choosing the arm with the largest mean.
Numerous extensions and variants of the classical stochastic MAB problem~\cite{lairob85} have been recently investigated.
These extensions are motivated by problems arising in various fields, including online recommendation systems (search engines, display ads, etc.), financial portfolio design~\cite{shen--15,huofu-17}, adaptive network routing~\cite{talebi18}, web crawling~\cite{kolobo20}, clinical trials~\cite{press-09,thomps33}, and, recently, in control theory~\cite{verstr20} and system identification~\cite{muller17,mulroj19}.

In this paper we consider a bandit problem in which, at every round, the agent can spread the available resources to many arms instead of allocating all the resources to only one arm.
This setup allows the agent to collect more information from the environment, since it perceives a pay-off from every arm where it has allocated resources, but at the cost of reducing the accuracy of the samples.
More specifically, the outcomes from each arm have a fixed mean for every experiment, but the variance of the samples is inversely proportional to the amount of resources allocated to each arm.
This poses an extra difficulty to the agent, which not only has to balance the exploration-exploitation trade-off, but now it also has to balance the precision invested in every sample.
The sequence of outcomes received by the agent at every round resembles the data collected in a traditional MAB along many experiments, where sampling a given arm many times emulates one noisy sample from its mean but with smaller variance.
By not allocating any resource to an arm, the variance of its sample diverges, which can be interpreted as missing or non-informative data, recovering the traditional MAB setup where the pay-offs are only revealed to the agent when an arm is selected.
The class of distributions addressed in this problem are Gaussian, two-dimensional, and where the minimum arm's variance (\ie, the one generating the data when all the resources are allocated to only one arm) is possibly unknown, leading to a 1-parameter or 2-parameter Gaussian bandit.
Gaussian distributions are particularly suitable for this class of problems, since their means and variances are decoupled, \ie, they can be fixed independently.

The bandit problem presented in this work is motivated by a fundamental problem in model-based control, where it is necessary to estimate the largest $\ell_2$-gain of the difference between a linear time-invariant dynamical system (the system we want to control) and its model, also known as the modeling error.
Estimating this quantity efficiently in a model-free fashion involves collecting input-output data from the modeling error when the input applied to the system produces an output of large gain (in Euclidean norm).
It is well known that the input signal achieving the largest gain is a sinusoid whose frequency is equal to the one at which the frequency response of the modeling error is the largest, also known as the peak frequency.
The problem of data collection can be addressed as a regret-minimization problem of choosing as input to the system a sinusoid of frequency different from the peak frequency.

\subsubsection*{Our contributions}
The main contributions of our work are:
\begin{enumerate}
\item the introduction of a new MAB problem called a bandit under weighted information, which allows resource-spreading policies;
\item a lower bound on the asymptotic regret incurred by a wide class of uniformly efficient policies;
\item a bandit algorithm, \WTS (Weighted Thompson Sampling), that can efficiently spread the resources at every round;
\item a thorough theoretical derivation of an upper bound on the asymptotic regret incurred by \WTS for the  2-parameter Gaussian bandit under weighted information, which matches the lower bound on the regret, thus establishing the asymptotic optimality of \WTS for these classes of problems; and
\item an application of \WTS to the gain estimation problem in control and system identification, where the goal is to estimate the $\Hinf$-norm of an unknown system from input-output data.
\end{enumerate}

The remainder of this paper is organized as follows: Section~\ref{sec:2_Problem_formulation} formalizes the MAB problem and its relation to linear bandits.
Section~\ref{sec:3_Lower_bound} introduces the notion of uniformly efficient policies and derives a lower bound for the regret of policies in this class.
Section~\ref{sec:4_WTS} presents \WTS together with its proof of optimality, while Section~\ref{sec:5_Applications} presents the application to the gain estimation problem in control and system identification.
Conclusions are presented in Section~\ref{sec:6_Conclusions}. 
For brevity, all the proofs appear in the supplementary material.

\subsection{Related work}

Fundamental limitations on the frequentist regret depend on the precise structure of the problem, and on the prior knowledge available to the agent, which was originally formalized by~\cite{burkat96} for a general class of problems.
The framework introduced in this work is very similar to the bandit problem with Gaussian rewards~\cite{latsze20}, which is one of the most fundamental stochastic MABs.
The first finite-time analysis for 1-parameter Gaussian bandits appeared in \cite{auer--02}, where \texttt{UCB-Normal} was introduced and proved to attain logarithmic regret.
Later, \texttt{kl-UCB} was introduced in~\cite{cappe-13}, where it was proved to match the lower bound of Lai and Robbins~\cite{lairob85}.
Regarding the 2-parameter Gaussian bandit,~\cite{aumusz09} introduced \texttt{UCB-V}, an algorithm that uses variance estimates to construct confidence intervals, which is shown to attain logarithmic but suboptimal regret.
Using a similar idea,~\cite{cowan-18} proved the optimality of their improved version named \texttt{ISM}.

Thompson Sampling (\TS)~\cite{thomps33} is a famous heuristic algorithm that was re-discovered in the last decade due to its excellent empirical performance~\cite{chali-11}.
The first thorough theoretical analysis of \TS was carried out in \cite{agrgoy12}, where a logarithmic (yet suboptimal) upper bound was derived for the regret incurred by \TS for bounded distributions.
A matching upper bound for \TS was derived in~\cite{kaufma12} for Bernoulli bandits, which was later extended in~\cite{korda-13} to 1-dimensional Gaussian bandits under a Jeffreys prior.
~\cite{hontak14} has developed a careful analysis for \TS under a 2-parameter (mean and variance) Gaussian bandit, concluding that its optimality crucially depends on the choice of the prior. 
In fact, the algorithm does not achieve optimality when, for example, a Jeffreys prior is employed, nor even a logarithmic asymptotic regret. 
Problem-independent upper bounds for \TS under bounded rewards can be found in~\cite{agrgoy12b}.

The idea of playing more than one arm per round given a certain budget is not new in machine learning.
Examples of problems where the budget is a stopping time appear in \cite{zhotom18,tran-t10}, however none of them considers weighted information being fed back to the agent: the feedback is the same as in a stochastic MAB that samples many arms per round.
The first 1-parameter Gaussian lower bound for a bandit with weighted information was presented in~\cite{muller17}, together with \WTS.
Later, \TS was proven to be asymptotically optimal for the 2-parameter Gaussian bandit, but when the outcomes are 2-dimensional and the regret is a nonlinear function of these outcomes~\cite{mulroj19}.

With respect to the application to gain estimation in control, some iterative approaches based on the power-iterations method of numerical linear algebra have been already proposed in the control community~\cite{wabahj10,rojas-12}, where the input signal is allowed to be designed in a sequential manner based on data from previous rounds. 
The specific problem of $\Hinf$-norm estimation has gained some attention in
computer science; for example, \cite{tubore18} has derived sharp asymptotic bounds on the error incurred by a method that firstly fits an FIR (finite impulse response) filter of $L$ coefficients to $N$-length data, in terms of $N$.

%% file: contents/2_Problem_formulation.tex
\label{sec:2_Problem_formulation}

Consider a stochastic multi-armed bandit problem (MAB) in which an agent can sample from $K$ different independent distributions $(\nu_k^t)_{k=1}^K$ at every round $t\in\N$.
The means of these distributions are unknown but fixed and denoted as $\bmu\eq (\bmu_1,\bmu_2,\dots,\bmu_K)$, where $\bmu_k\in\R^2$, for each $k=1,\dots,K$; however, unlike in a standard stochastic MAB setup, the variance of these distributions will depend on the action taken by the agent.
We define $k^\star \eq \arg\min_{k=1,\dots,K}\norm{\bmu_k}$, $\bmu^\star \eq \bmu_{k^{\star}}$, and the gaps $\Delta_k\eq \norm{\bmu^\star}-\norm{\bmu_k}$, where $\norm{\cdot}$ is the Euclidean norm.
At every round $t\in\N$, the agent chooses a \emph{power profile}\footnote{The dependence of the power profiles on $\bpi$ is made explicit in the notation, however, we might drop the superscript when this dependence is clear.}
$p^{t,\bpi}\eq(p^{t,\bpi}_1,p^{t,\bpi}_2,\dots,p^{t,\bpi}_K)$ according to policy $\bpi$, and it receives a random outcome tuple $X^t\eq(X^t_1,X^t_2,\dots,X^t_K)$, where $X^t_k \in \R^2$ satisfies
\begin{align}
X^t_k \sim \Nsp\left(\bmu_k,\frac{\sigma^2_k}{2p^{t,\bpi}_k}\bI_2\right)
\label{eq:X}
\end{align}
for a fixed and unknown vector of variances $\bsigma \eq (\sigma^2_1,\sigma^2_2,\dots,\sigma^2_K)$, where $\bI_2$ denotes the identity matrix in $\R^{2\times 2}$.
In other words, the Gaussian outcome distribution $\nu^t_k$ is parametrized by the amount of power $p^{t,\bpi}_k$ allocated to that particular arm, which only affects the variance of the distribution.
The power profiles are restricted to fulfil $p^{t,\bpi}\in\Lambda$ at every round, where $\Lambda$ is the simplex $\Lambda\eq\{\xi\in[0,1]^K\colon \sum_{k=1}^K \xi_k = 1\}$.
The selection of $\bpi$ may be adaptive and depend on the power levels and observed outcomes in previous experiments.
Let $\bPi$ denote the set of policies $\bpi$ such that the power profile $p^{t,\bpi}$ assigned under policy $\bpi$ at experiment $t$ is $\Fsp_t$-measurable, where $\Fsp_t$ is the sigma-algebra generated by the previous observations $(p^{1,\bpi},X^1,p^{2,\bpi},X^2,\dots,p^{t-1,\bpi},X^{t-1})$.

We are interested on finding a policy $\bpi\in\bPi$ minimizing the frequentist regret of not allocating all the power at arm $k^\star$ (having the largest mean in magnitude).
To this end, we introduce the following notion of expected cumulative regret:
\begin{align}
\mean{R^{\bpi}_{\bmu,\bsigma}(T)}
=
\sum_{t=1}^T\sum_{k=1}^K \norm{\bmu_k} (p^{k,\star}_k - \mean{p^{t,\bpi}_k}),
\label{eq:regret}
\end{align}
where $p^{t,\star}$ is the power profile assigned by an oracle policy that knows $k^\star$ in hindsight, \ie, assigning $p^{t,\star}_{k^\star} = 1$ and $p^{t,\star}_k=0$, for every suboptimal arm $k\neq k^\star$, at every round $t\in\N$.
Observe that the dependence of the regret on $\bmu$ and $\bsigma$ is made explicit.
We formalize this problem as the regret minimization (RM) problem:
\begin{align}
\mbox{(RM)} \quad \min_{\bpi\in\bPi} \mean{R_{\bmu,\bsigma}(T)}.
\notag
\end{align}

For the sake of comparison, we introduce a sub-class of policies in $\bPi$ that sample only one arm $k_t$ at each round $t\in\N$, by applying the power profile $p^{t,\bpi}_{k_t}=1$ and $p^{t,\bpi}_k=0$ for every other arm $k\neq k_t$.
This class is denoted as $\bPi_{\text{NS}}\subset \bPi$, where NS stands for non-spreading, in the sense that our problem considers bandits that can ``spread" the available budget to many arms per round.
Certainly, algorithms in $\bPi_{\text{NS}}$ correspond to traditional MAB policies where only one arm per round is sampled.

\subsection{Relation to linear bandits}
The expected cumulative regret in~\eqref{eq:regret} resembles the one measuring the performance of a linear bandit, \ie, where the reward is a noisy linear function of the distribution means~\cite{latsze20}.
To see this, define the reward perceived by the agent at every round $t\in\N$ as  the linear combination
\begin{align}
\sum_{k=1}^K p^{t,\bpi}_k \left(\norm{X^t_k}^2 - \sigma_k^2\right),
\label{eq:reward}
\end{align}
where we subtract the variance of the outcomes because we are not interested in that part of the (expected) reward generated by the variance of the outcomes. 
The agent then aims to maximize the expected cumulative reward
\begin{align}
\mean{
\sum_{t=1}^T \sum_{k=1}^K p^{t,\bpi}_k \left(\norm{X^t_k}^2-\sigma_k^2\right)
}
=
\sum_{t=1}^T \sum_{k=1}^K \mean{p^{t,\bpi}_k}\norm{\bmu_k}^2 ,
\end{align}
which is maximized by the oracle policy.
Since such an optimal policy is independent of the exponent in $\norm{\bmu_k}^2$, one can instead consider $\max_{\bpi\in\bPi} \sum_{t=1}^T \means{p^{t,\bpi}_k}\norm{\bmu_k}$, which is equivalent to solving~(RM).

The difference between solving~(RM) and a traditional linear bandit lies in the more detailed feedback we receive in~(RM) at every round.
More precisely, the agent measures each of the outcomes in the linear combination~\eqref{eq:reward} which provides information about all distributions $(\nu_k^t)_k$ simultaneously.
As shown in Section~\ref{sec:3_Lower_bound}, this crucial difference allows us to dramatically reduce the rate of growth of the expected cumulative regret $\mean{R^{\bpi}_{\bmu,\bsigma}(T)}$ in~\eqref{eq:regret} from $\Theta(\sqrt{T})$~\cite{dani--08,latsze20} to $\Theta(\log T)$.
The problem addressed in this work thus presents a new instance of a stochastic MAB that lies between linear bandits and traditional MABs, since the cost function is linear in the actions but the information received by the agent resembles the one collected in a traditional MAB after many rounds.

%% file: contents/3_Lower_bound.tex
\label{sec:3_Lower_bound}

In this section we introduce a class of algorithms that formally address the exploration-exploitation dilemma successfully enough in order to quickly discriminate the optimal arm.
Following the notion originally introduced by~\cite{lairob85} for traditional MABs, we call this family the set of \emph{uniformly efficient policies}.
In the context of bandits with weighted information, this notion has already been introduced in~\cite{muller17}, and it corresponds to the spreading version of the one in~\cite{lairob85}.

\begin{definition}[Uniform efficiency]
\label{def:uniform}
A policy $\bpi$ is said to be \emph{uniformly efficient} if the cumulative power $p^{t,\bpi}_k$ allocated at every suboptimal arm $k\neq k^\star$ satisfies $\means{\sum_{t=1}^T p^{t,\bpi}_k} = \o(T^{\alpha})$, for every $\alpha > 0$.
The set of uniformly efficient strategies is denoted as $\bPi^\star$.
The subset of non-spreading but uniformly efficient policies in $\bPiNS$ is denoted as $\bPiNS^\star$.
\end{definition}

In their seminal paper~\cite{lairob85}, Lai and Robbins derived a tight asymptotic (in $T$) regret lower bound satisfied by every non-spreading uniformly efficient policy.
Following this idea,~\cite{mulroj19} derived a lower bound for the regret in~\eqref{eq:regret}, for policies in $\bPiNS^\star$, showing that 
$\liminf_{T\to\infty}\means{R^{\bpi}_{\bmu,\bsigma}(T)}/\log T = \sum_{k\neq k^\star} \sigma_k^2/\Delta_k$ when $\bsigma$ is known, and that 
$\liminf_{T\to\infty}\means{R^{\bpi}_{\bmu,\bsigma}(T)}/\log T = \sum_{k\neq k^\star} \Delta_k/\log(1+\Delta_k^2/\sigma_k^2)$ when $\bsigma$ is unknown.
This result suggests that it becomes easier for the agent to identify suboptimal arms when the variance of the noise in the outcomes is known in hindsight.
For bandits under weighted information,~\cite{muller17} had already reported that $\liminf_{T\to\infty}\means{R^{\bpi}_{\bmu,\bsigma}(T)}/\log T = \sum_{k\neq k^\star} \sigma_k^2/\Delta_k$ for every $\bpi\in\bPi^\star$ when $\bsigma$ is known, suggesting that there may not be any performance improvement (asymptotically) when spreading strategies are preferred.
The following result describes the missing lower bound: when $\bsigma$ is unknown, one can recover the same lower bound when resources can be spread among arms, as if $\bsigma$ were revealed to the agent in hindsight.

\begin{theorem}[Regret lower bound]
\label{thm:lower_bound}
Consider~\emph{(RM)} with unknown parameters $\bmu$ and $\bsigma=(\sigma_1^2,\sigma_2^2,\dots,\sigma_K^2)$.
Under every uniformly efficient policy $\bpi\in\bPi^\star$, the expected cumulative regret, as defined in~\eqref{eq:regret}, satisfies
\begin{align}
\liminf_{T\to\infty}\frac{\mean{R^{\bpi}_{\bmu,\bsigma}(T)}}{\log T} 
\geq 
\sum_{k\neq k^\star} \frac{\sigma_k^2}{\Delta_k^2}.
\end{align}
\end{theorem}

The result in Theorem~\ref{thm:lower_bound} completes the set of lower bounds for policies in $(\bPi^\star,\bPiNS^\star)$ when $\bsigma$ is known or not to the agent.
It also formalizes the idea that sampling many arms simultaneously at every round can lead to better performance, in terms of cumulative expected regret, when the variances in $\bsigma$ are unknown.
Table~\ref{table:lower_bounds} puts this result in perspective by showing the different regret lower bounds under known/unknown variance for uniformly efficient policies in the set of spreading and non-spreading strategies $\bPiNS^\star$ and $\bPi^\star$, respectively.
The lower-right entry, originally unknown, is filled by Theorem~\ref{thm:lower_bound}.

\begin{table}
  \caption{Regret lower bounds.}
  \label{table:lower_bounds}
  \centering
  \begin{tabular}{ccc}
    \toprule
    \multicolumn{3}{c}{
    ${\displaystyle\liminf_{T\to\infty}
    \frac{\mean{\regret^{\bpi}_{\bmu,\bsigma}(T)}}{\log T}\geq}$}                   	\\
    \cmidrule(r){2-3}
      & Known $\bsigma=(\sigma_1^2,\sigma_2^2,\dots,\sigma_K^2)$ 
                  & Unknown $\bsigma=(\sigma_1^2,\sigma_2^2,\dots,\sigma_K^2)$                                               	\\
    \midrule
    $\bpi\in\bPiNS^\star$ 
    	& ${\displaystyle \sum_{k\neq k^\star} 
    			\frac{\sigma^2_k}{\Delta_k}}$~\cite{bubces12} 
    		& ${\displaystyle\sum_{k\neq k^\star} 
    			\frac{\Delta_k}{\log\left(1+\frac{\Delta_k^2}{\sigma_k^2
    					}\right)}}$~\cite{mulroj19}
    \\
    $\bpi\in\bPi^\star$
    	& ${\displaystyle \sum_{k\neq k^\star} 
    			\frac{\sigma^2_k}{\Delta_k}}$~\cite{muller17}   
    	& ${\displaystyle \sum_{k\neq k^\star} \frac{\sigma^2_k}{\Delta_k}}$ \\
    \bottomrule
  \end{tabular}
\end{table}

%% file: contents/4_WTS.tex
\label{sec:4_WTS}
Weighted Thompson Sampling (\WTS) is a spreading strategy in $\bPi$ originally introduced in~\cite{muller17}.
This algorithm is based on the algorithm by Thompson~\cite{thomps33}, called Thompson Sampling (\TS), and it corresponds to a Bayesian policy that condenses gathered information in the form of posterior distributions of each arm being optimal but, instead of randomizing the action (as \TS does), \WTS uses this posterior distribution as a power profile $p^t$.
In this section we develop a thorough analysis of \WTS which overlaps with some of the ideas used to study \TS for the solution of (RM) in~\cite{mulroj19}.

Just as with \TS, \WTS starts with a prior distribution $\rho^1 = (\rho^1_k)_{k=1}^K$ encoding our confidence on each arm being the optimal arm, \ie, $\rho^1_k = $Prob$\{\tk_1 = k\given\Fsp_1\}$, where $\tk_t$ is a random variable whose distribution (conditioned to $\Fsp_t$) captures our confidence of $k$ being optimal given the history, and $\Fsp_1=\{\phi,\Omega\}$ (\ie, no data).
Recall that $\Fsp_t$ denotes the sigma algebra generated by $(p^{1,\bpi},X^1,p^{2,\bpi},X^2,\dots,p^{t-1,\bpi}$ $X^{t-1})$ under policy $\bpi\in\bPi$.
At every round $t\in\N$, \WTS applies the power profile $p^{t,\WTS}=\rho^t$ and then updates its belief of each arm's optimality $\rho^{t+1}$ based on the history up to round $t$.
Obtaining $\rho^t$ in closed form is, in general, not possible.
This can be overcome by introducing the posterior means $\tbmu_k(t)\sim f_{\bmu_k\given\Fsp_t}$, where $f_{\bmu_k\given\Fsp_t}$ is the posterior distribution on the mean $\bmu_k$ given the data up to round $t-1$.
As discussed in~\cite{kaufma14}, if $\norm{\tbmu^\star(t)}\eq\max_k\norm{\tbmu_k(t)}$, then
\begin{align}
\rho^t_k 
\eq
\text{P}\{\tk_t=k\given\Fsp_t\}
= 
\means{\onesmall{\tk_t=k}\given\Fsp_t}
=
\prob{\norm{\tbmu^\star(t)}=\norm{\tbmu_k(t)}\given\Fsp_t},
\label{eq:rho_wts}
\end{align}
which means that $\rho^t$ is completely determined by $(f_{\bmu_1\given\Fsp_t}, f_{\bmu_2\given\Fsp_t},\dots, f_{\bmu_K\given\Fsp_t})$.
It is important to mention that finding a closed form for the mapping from $(f_{\bmu_1\given\Fsp_t}, f_{\bmu_2\given\Fsp_t},\dots, f_{\bmu_K\given\Fsp_t})$ to $\rho^t$ involves the calculation of very complicated $K$-dimensional integrals, that can however be approximated by Monte Carlo calculations~\cite{fishma96}, as we explain in Appendix~\ref{app:rho}.

It is clear that a key component in \WTS is to be able to obtain $f_{\bmu_k\given\Fsp_t}$ at every round $t\in\N$, and a method to achieve this task involves computing sufficient statistics for the distribution of the outcomes.
To obtain $f_{\bmu_k\given\Fsp_t}$ we select a prior distribution representing our total ignorance on what the real value of $(\bmu,\bsigma)$ is by considering a uniform improper distribution $f_{\bmu_k,\sigma^2_k\given\Fsp_1}(\bm,\zeta^2)\propto 1$, for every $(\bm,\zeta^2)\in\R^2\times (0,\infty)$ and for every $k=1,\dots,K$.
The posterior $f_{\bmu_k\given\Fsp_t}(\bm)$ is then obtained as $\int_0^\infty f_{\bmu_k,\sigma_k^2}(\bm,\zeta^2) d\zeta^2$.
This paper shows, in Section~\ref{subsec:optimality_WTs}, that \WTS is optimal under such a prior, even though other choices of priors may also attain optimality~\cite{hontak14}. 

\subsection{Sufficient Statistics}
Here we show that the sample mean, the sample variance and the trajectory of the power profiles $(p^{1,\bpi},p^{2,\bpi},\dots,p^{t,\bpi})$  are sufficient statistics for the data collected up to round $t$.
We additionally characterize the distribution of the sample mean, the sample variance, and the posterior mean distribution $f_{\bmu_k\given\Fsp_t}$. These results are presented in the following lemma.

\begin{lemma}
\label{lem:WTS_suff}
For each arm $k\in\setK$,
\begin{align}
\xx_k(t) \eq \frac{\sum_{\ell=1}^t p^\ell_k X^\ell_k }{\sum_{\ell=1}^t p^\ell_k},
\qquad
S_k(t) \eq \sum_{\ell=1}^t p^\ell_k\norm{X^\ell_k - \xx_k(t)}^2
\label{eq:xxS}
\end{align}
are sufficient statistics for $(X^\ell_k)_{\ell=1}^t$ conditioned on $p^1_k,p^2_k,\dots,p^t_k$.
Furthermore, conditioned on the trajectory of the power profiles, $\xx_k(t)\sim\Nsp(\bmu_k,\sigma_k^2\bI_2/(2\sum_{\ell=1}^t p^\ell_k))$ and $2S_k(t)/\sigma_k^2\sim\chi^2_{2(t-1)}$ are statistically independent.
\end{lemma}


The following result states that the posterior means follow a bivariate t-student distribution~\cite{papoul91} that is symmetrically distributed around the empirical mean $\xx_k(t)$ with variance proportional to $S_k(t)/t$, at every round $t\in\N$ and for every arm $k=1,\dots,K$.

\begin{lemma}
\label{lem:post_WTS}
Consider the improper uniform prior distribution $f_{\bmu_k,\sigma_k^2\given\Fsp_1}\propto 1$.
Then, the posterior distribution $f_{\bmu_k\given\Fsp_t}$ at round $t\geq 4$ is
\begin{align}
f_{\bmu_k\given\Fsp_t}(\tbmu)
&=
\frac{\sum_{\ell=1}^{t-1}p^\ell_k(t-3)}{\pi S_k(t-1)}\left( 1 + \frac{\sum_{\ell=1}^{t-1}p^\ell_k\norm{\tbmu-\xx_k(t-1)}^2}{S_k(t-1)}\right)^{-t+2},
\label{eq:post_wts}
\end{align}
for every $k=1,\dots,K$.
\end{lemma}

We can now formalize the \WTS algorithm introduced at the beginning of this section by using Lemmas~\ref{lem:WTS_suff} and~\ref{lem:post_WTS} to update $\rho^t$ in a tractable fashion.
In fact, having have a closed-form expression for the posterior means in~\eqref{eq:post_wts} allows us to approximate $\rho^t$ by means of~\eqref{eq:rho_wts} and Monte Carlo simulations, as explained in Appendix~\ref{app:rho}.
We summarize \WTS in  Algorithm~\ref{alg:wts}, where we remark that, by Lemma~\ref{lem:post_WTS}, each arm needs to be sampled at least 3 times.

\begin{algorithm}
\caption{\quad \WTS: proposed implementation via sufficient statistics}
\label{alg:wts}
\begin{algorithmic}[1]
      	\State Input: $T$, $\rho^1 = (1/K, 1/K,\dots,1/K)$ (prior distribution for $\tk_1$)
        \For{$t = 1$ to $3$}
            \State Apply the power profile $p^{t,\WTS} = \rho^1$ and collect the outcome $X^t$
            \State Update the sufficient statistics $\xx_k(t)$ and $S_k(t)$ according to Lemma~\ref{lem:WTS_suff}
        \EndFor
		\State Obtain $(f_{\bmu_1\given\Fsp_4},f_{\bmu_2\given\Fsp_4},\dots,f_{\bmu_K\given\Fsp_4})$ via Lemma~\ref{lem:post_WTS} and compute $\rho^3$
		\For{$t = 4$ to $T$}
            \State Apply the power profile $p^{t,\WTS} = \rho^1$ and collect the outcome $X^t$
            \State Update the sufficient statistics $\xx_k(t)$ and $S_k(t)$ according to Lemma~\ref{lem:WTS_suff}
            \State Obtain $(f_{\bmu_1\given\Fsp_{t+1}},f_{\bmu_2\given\Fsp_{t+1}},\dots,f_{\bmu_K\given\Fsp_{t+1}})$ via Lemma~\ref{lem:post_WTS} and compute $\rho^{t+1}$
        \EndFor
\end{algorithmic}
\end{algorithm}


\subsection{Optimality of \WTS} \label{subsec:optimality_WTs}
Here we provide a theoretical analysis of \WTS, showing that its upper bound matches the regret lower bound predicted by Theorem~\ref{thm:lower_bound}.
\begin{theorem}[Upper bound]
\label{thm:upper_WTS_unknown}
Consider~\emph{(RM)} under unknown $(\bmu,\bsigma)$, and the improper uniform prior distribution $f_{\bmu_k,\sigma_k^2\given\Fsp_1}\propto 1$.
Then, the regret, as defined in~\eqref{eq:regret}, incurred by {\normalfont \WTS} satisfies
\begin{align}
\limsup_{T\to\infty} \frac{\mean{\regret^{\textnormal{WTS}}_{\bmu,\bsigma}(T)}}{\log T} \leq \sum_{k\neq k^\star} \frac{\sigma_k^2}{\Delta_k}.
\end{align}
\end{theorem}

The reason why \WTS is able to attain the same performance regardless its knowledge of $\bsigma$ follows from the fact that, for every $k=1,\dots,K$, the outcomes from arm $k$ provide the same information about $\sigma_k^2$ despite how small $p^t_k$ is at round $t$.
In fact, the sample variance $2 S_k(t)/t$ concentrates exponentially around $\sigma_k^2$ (much faster than the sample means $\xx_k(t)$), meaning that \WTS can quickly make decisions as if it knew the value of $\bsigma$ in hindsight.

\subsection{Simulation study 1: regret of \WTS and \TS under known/unknown variance}
\label{subsec:WTS_sim}
\begin{figure}[t]
\centering 
\def\svgwidth{1\textwidth}
\small
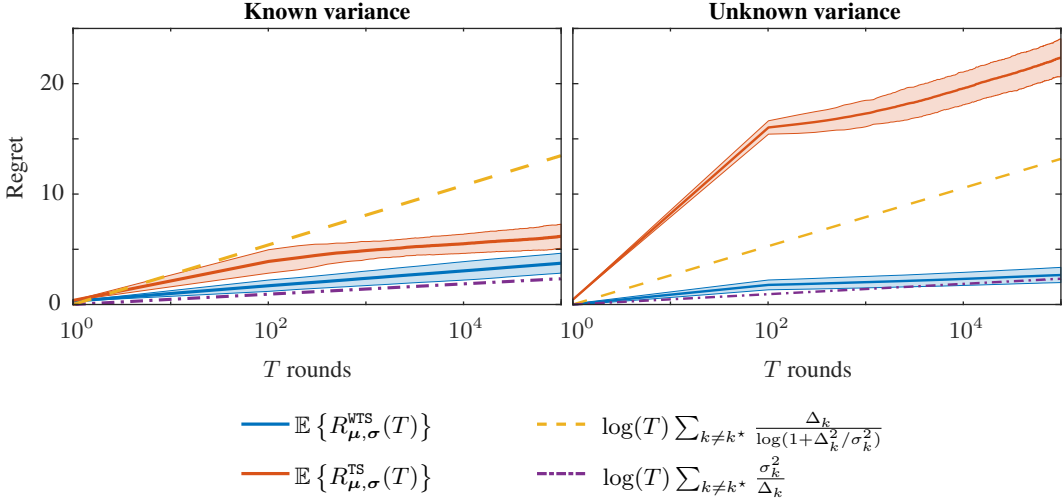 
\caption{Summary of the performances attained by \WTS and \TS, in the light of the lower bounds of Table~\ref{table:lower_bounds}, under known and unknown $\bsigma$.
The shaded areas represent 2 standard deviations.
The implementation of \WTS and \TS under known $\bsigma$ is described in Appendix~\ref{app:simulation}.
}
\label{fig:all_algs}
\end{figure}

We consider an arbitrary system with prefixed values for $\bmu,\bsigma,K$, on which we test \TS and \WTS for known and unknown $\bsigma$.
The values of $\bmu,\bsigma,K$ are chosen so as to highlight the different lower bounds in Table~\ref{table:lower_bounds}, and their values are detailed in Appendix~\ref{app:simulation}.
We run 300 Monte Carlo simulations to approximate the expected cumulative regret for each of them during $T=10^5$ rounds.
In line with the results derived in~\cite{muller17,mulroj19}, the simulations, depicted in Figure~\ref{fig:all_algs}, show that \TS and \WTS are matching algorithms for each of the two setups (known and unknown $\bsigma)$, \ie, their asymptotic expected cumulative regret matches each of the lower bounds summarized in Table~\ref{table:lower_bounds}.
When $\bsigma$ is known (left), the asymptotic expected regret is the same for \TS (red) and \WTS (blue), meaning that there is no improvement, asymptotically, in spreading the resources at every round.
Both algorithms match the same lower bound (purple) prescribed for both $\bPi^\star$ and $\bPiNS^\star$ under known $\bsigma$~\cite{muller17}.
When the noise is unknown (right), such lower bounds are different, where we can observe that \TS (red) and \WTS (blue) match their respective lower bounds (yellow and purple, respectively).

\normalfont

%% file: figures/all_algs2.pdf_tex
\begingroup%
  \makeatletter%
  \providecommand\color[2][]{%
    \errmessage{(Inkscape) Color is used for the text in Inkscape, but the package 'color.sty' is not loaded}%
    \renewcommand\color[2][]{}%
  }%
  \providecommand\transparent[1]{%
    \errmessage{(Inkscape) Transparency is used (non-zero) for the text in Inkscape, but the package 'transparent.sty' is not loaded}%
    \renewcommand\transparent[1]{}%
  }%
  \providecommand\rotatebox[2]{#2}%
  \newcommand*\fsize{\dimexpr\f@size pt\relax}%
  \newcommand*\lineheight[1]{\fontsize{\fsize}{#1\fsize}\selectfont}%
  \ifx\svgwidth\undefined%
    \setlength{\unitlength}{200bp}%
    \ifx\svgscale\undefined%
      \relax%
    \else%
      \setlength{\unitlength}{\unitlength * \real{\svgscale}}%
    \fi%
  \else%
    \setlength{\unitlength}{\svgwidth}%
  \fi%
  \global\let\svgwidth\undefined%
  \global\let\svgscale\undefined%
  \makeatother%
  \begin{picture}(1,0.475)%
    \lineheight{1}%
    \setlength\tabcolsep{0pt}%
    \put(0.52267855,0.14550365){\color[rgb]{0.14901961,0.14901961,0.14901961}\makebox(0,0)[lt]{\lineheight{1.25}\smash{\begin{tabular}[t]{l}$10^0$\end{tabular}}}}%
    \put(0.70770498,0.14550365){\color[rgb]{0.14901961,0.14901961,0.14901961}\makebox(0,0)[lt]{\lineheight{1.25}\smash{\begin{tabular}[t]{l}$10^2$\end{tabular}}}}%
    \put(0.89273141,0.14550365){\color[rgb]{0.14901961,0.14901961,0.14901961}\makebox(0,0)[lt]{\lineheight{1.25}\smash{\begin{tabular}[t]{l}$10^4$\end{tabular}}}}%
    \put(0.71671912,0.10790209){\color[rgb]{0.14901961,0.14901961,0.14901961}\makebox(0,0)[lt]{\lineheight{1.25}\smash{\begin{tabular}[t]{l}$T$ rounds\end{tabular}}}}%
    \put(0.03628919,0.17196598){\color[rgb]{0.14901961,0.14901961,0.14901961}\makebox(0,0)[lt]{\lineheight{1.25}\smash{\begin{tabular}[t]{l}0\end{tabular}}}}%
    \put(0.02824457,0.27677598){\color[rgb]{0.14901961,0.14901961,0.14901961}\makebox(0,0)[lt]{\lineheight{1.25}\smash{\begin{tabular}[t]{l}10\end{tabular}}}}%
    \put(0.02824457,0.38158598){\color[rgb]{0.14901961,0.14901961,0.14901961}\makebox(0,0)[lt]{\lineheight{1.25}\smash{\begin{tabular}[t]{l}20\end{tabular}}}}%
    \put(0.01141385,0.27734221){\color[rgb]{0.14901961,0.14901961,0.14901961}\rotatebox{90.00000248}{\makebox(0,0)[lt]{\lineheight{1.25}\smash{\begin{tabular}[t]{l}Regret\end{tabular}}}}}%
    \put(0,0){\includegraphics[width=\unitlength,page=1]{all_algs2.pdf}}%
    \put(0.56354918,0.0054576){\color[rgb]{0,0,0}\makebox(0,0)[lt]{\lineheight{1.25}\smash{\begin{tabular}[t]{l}$\log(T)\sum _{k\neq k^\star}\frac{\sigma_k^2}{\Delta_k}$\end{tabular}}}}%
    \put(0.56258643,0.05358348){\color[rgb]{0,0,0}\makebox(0,0)[lt]{\lineheight{1.25}\smash{\begin{tabular}[t]{l}$\log(T)\sum_{k\neq k^\star}\frac{\Delta_k}{\log(1+\Delta_k^2/\sigma_k^2)}$\end{tabular}}}}%
    \put(0.27051486,0.05443983){\color[rgb]{0,0,0}\makebox(0,0)[lt]{\lineheight{1.25}\smash{\begin{tabular}[t]{l}$\mean{\regret^{\WTS}_{\bmu,\bsigma}(T)}$\end{tabular}}}}%
    \put(0.27051486,0.00631395){\color[rgb]{0,0,0}\makebox(0,0)[lt]{\lineheight{1.25}\smash{\begin{tabular}[t]{l}$\mean{\regret^{\TS}_{\bmu,\bsigma}(T)}$\end{tabular}}}}%
    \put(0.04646984,0.1455419){\color[rgb]{0.14901961,0.14901961,0.14901961}\makebox(0,0)[lt]{\lineheight{1.25}\smash{\begin{tabular}[t]{l}$10^0$\end{tabular}}}}%
    \put(0.23149627,0.1455419){\color[rgb]{0.14901961,0.14901961,0.14901961}\makebox(0,0)[lt]{\lineheight{1.25}\smash{\begin{tabular}[t]{l}$10^2$\end{tabular}}}}%
    \put(0.4165227,0.1455419){\color[rgb]{0.14901961,0.14901961,0.14901961}\makebox(0,0)[lt]{\lineheight{1.25}\smash{\begin{tabular}[t]{l}$10^4$\end{tabular}}}}%
    \put(0.24051041,0.10794034){\color[rgb]{0.14901961,0.14901961,0.14901961}\makebox(0,0)[lt]{\lineheight{1.25}\smash{\begin{tabular}[t]{l}$T$ rounds\end{tabular}}}}%
    \put(0.22192197,0.44880113){\color[rgb]{0,0,0}\makebox(0,0)[lt]{\lineheight{1.25}\smash{\begin{tabular}[t]{l}\textbf{Known variance}\end{tabular}}}}%
    \put(0.66442198,0.44880113){\color[rgb]{0,0,0}\makebox(0,0)[lt]{\lineheight{1.25}\smash{\begin{tabular}[t]{l}\textbf{Unknown variance}\end{tabular}}}}%
    \put(0,0){\includegraphics[width=\unitlength,page=2]{all_algs2.pdf}}%
  \end{picture}%
\endgroup%

%% file: contents/5_Applications.tex
\label{sec:5_Applications}

\begin{figure}[t]
\centering 
\def\svgwidth{0.348\columnwidth} 
\small
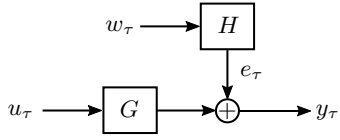 
\caption{Mathematical model of a linear system with additive non-white noise.}
\label{fig:system}
\end{figure}
\normalfont

In this section we study the extension of \WTS to an important problem in control and system identification~\cite{zhodoy96,ljung-99}.
One of the fundamental steps in designing a model-based controller for a dynamical linear and time-invariant (LTI) system is to estimate the mismatch between the real system and its LTI model in terms of the largest gain it can induce, also known as the induced $\ell_2$-gain or $\Hinf$-norm of the system~\cite{vdscha17}.
The system depicted in Figure~\ref{fig:system} is suitable for studying the problem of estimating such a quantity from input-output data.
To see this, let $G_0$ and $\hat{G}$ denote a system and its model, respectively, and define $G=G_0-\hat{G}$.
Then measurements from $G$ can be collected by exciting the real system $G_0$ and its model $\hat{G}$ with the same input and then subtracting their respective outputs.

Following the setup described in~\cite{muller17,muller18}, let $G(\ejw)$ and $H(\ejw)$ denote the transfer functions of the unknown, causal, stable LTI systems $G$ and $H$ in Fig.~\ref{fig:system}, where $w$ is Gaussian $\Nsp(0,1)$ white noise, and $j\eq\sqrt{-1}$.
We are interested on estimating the $\Hinf$-norm of $G$,defined as $\beta\eq\norm{G}_{\infty} = \max_{\w\in[0,\pi]} |G(\ejw)|$.
It is well known~\cite{kay---93} that estimating $\beta$ from input-output data crucially depends on the input signal.
In fact, applying a sinusoidal of frequency equal to $\arg\max_{\w\in[0,\pi]} |G(\ejw)|$ provides the best possible data for this task because it maximizes the ratio $\norm{Gu}/\norm{u}$~\cite{fairma98,kailat80}.
In line with this argument, we study the gain estimation problem as two sub-problems: the one of generating the input signal and that of using the collected data in a point-estimator of $\beta$.
The data is collected in a sequence of independent experiments\footnote{Statistically independent experiments can be achieved by either waiting long enough between experiments (so that the natural response of $G$ due to initial conditions decays exponentially to zero), by using a controller to bring the state of the system to zero, or, if possible, by manually resetting $G$.}, where at each round $t\in\N$, the agent is allowed to design the input signal $u^t=(u^t_\tau)_{\tau=0}^{N-1}$ and collect the output $y^t=(y^t_\tau)_{\tau=0}^{N-1}$, for some prefixed $N$.

We delegate the problem of optimally generating the input signal $u^t$ in an adaptive fashion to \WTS.
Since $G$ is LTI, we restrict the class of input signals to a sum of sinusoids (\ie, a multisine) of $K$ predetermined frequencies $(2\pi k/(2K+1))_{k=1}^K$ which discretize the frequency range $[0,\pi]$ into $K$ equispaced frequencies, where the agent can choose the amplitude of each sinusoid.
This can be achieved by choosing $N=2K+1$ and then letting the agent design $u^t$ in the frequency domain.
More specifically, if $U^t(j\w) \eq \sum_{\tau=0}^{N-1}u^t_\tau \exp{-j\w\tau}$ denotes the discrete Fourier transform~\cite{oppsch99} of $u^t$, then the agent designs $ |U^t(j\wk)|^2 =  p^t_k$ at the frequencies of interest $\wk =2\pi k/(2K+1)$.
By applying an input $u^t$ satisfying\footnote{This can be achieved by, for example, taking the inverse Fourier transform of the sequence $(0,\sqrt{p^t_1},\sqrt{p^t_2},\dots,\sqrt{p^t_K},\sqrt{p^t_K},\dots,\sqrt{p^t_2},\sqrt{p^t_1}$), or by setting $u^t = (u^t_\tau)_{\tau=1}^N =  \sum_{k=1}^K \sqrt{p^t_k} \sin(\wk \tau)$.} $|U^t(j\wk)|^2=p^t_k$, the input output data $(u^t,y^t)$ satisfies
\begin{align}
\frac{Y^t(j\wk)}{U^t(j\wk)} = G(\ejwk) + \frac{E^t(j\wk)}{U^t(j\wk)},
\end{align}
where $E^t(j\wk)$ is a circularly symmetric complex zero-mean and white sequence whose real and imaginary parts are independent for every $k=1,\dots,K$ and $t\in\N$~\cite{aguero10}.
Moreover, its real and imaginary parts are Gaussian with variance $|H(\ejwk)|/2$.
The $(2K+1)$-length noisy output is completely revealed to the agent after every experiment, who can then define the sequence of outcomes $X^t_k \eq [\Re\{Y^t(j\wk)/U^t(j\wk)\} \quad \Im\{Y^t(j\wk)/U^t(j\wk)\}]^\top$, $k=1,\dots,K$.
In consequence, $X^t_k\sim\Nsp(\bmu_k,\sigma^2_k/(2p^t_k)\bI_2)$, where $\bmu_k = [\Re G(\ejwk)\quad \Im G(\ejwk)]^\top$ and $\sigma_k^2 \eq |H(\ejwk)|^2$, which recovers the bandit problem under weighted information defined in Section~\ref{sec:2_Problem_formulation}.
The goal of the agent collecting the data is then to find the power profile $p^t$ that maximizes $\norm{Gu}$ in the class of multisine inputs which, for $K$ large enough, reasonably approximates $\norm{G}_\infty\approx\max_{k=1,\dots,K} |G(\ejwk)|$.

The point-estimator for $\beta=\norm{G}_\infty$ is independent of the underlying  data-collection algorithm and, for the sake of simplicity, it is chosen as 
$\hat{\beta}_t \eq \norm{\xx_{\hat{k}_t}(t)}$, with $\hat{k}_t = \arg\max_{k=1,\dots,K} \sum_{\ell=1}^t p^{\ell,\WTS}_k$.


\subsection{Simulation study 2: \WTS in the gain estimation problem}
\label{subsec:simulation_mse}
\begin{figure}[t]
\small
\centering 
\def\svgwidth{1\columnwidth} 
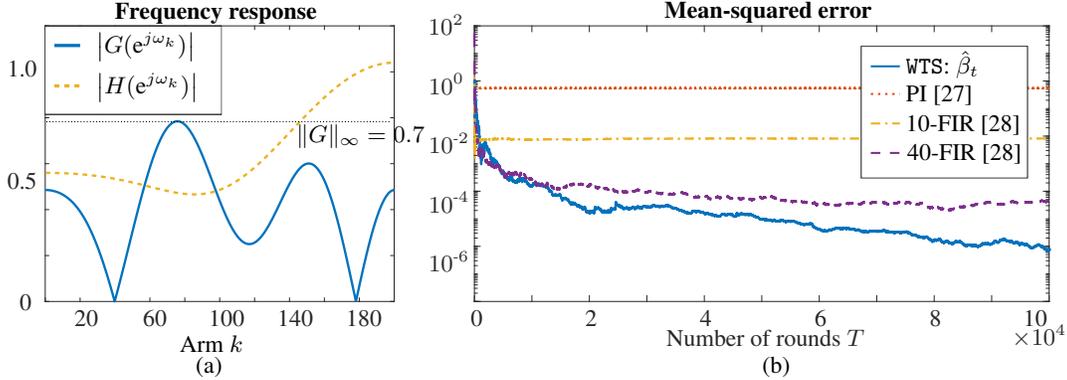 
\caption{Simulation study in Section~\ref{subsec:simulation_mse}.
The left-hand figure (a) depicts the frequency response of filters $G$ and $H$ in Fig.~\ref{fig:system}, while the right-hand figure (b) shows the performance attained by different state-of-the-art estimators and our proposed method.
}
\label{fig:double}
\end{figure}
\normalfont

Consider the configuration shown in Fig.~\ref{fig:system}, where $G$ and $H$ have the frequency responses depicted in Fig.~\ref{fig:double}a.
This simulation study is particularly confusing for any $\Hinf$-norm estimation algorithm since the frequency response of $G$ is almost completely covered by the one of $H$.
We implement \WTS with $K=200$ arms (to avoid large discretization errors) with an estimator $\hat{\beta}_t$, together with the power iterations (\texttt{PI}) method~\cite{wabahj10,rojas-12}, and a model-based approach that first derives an FIR model of lengths 10 (\texttt{10-FIR}) and 40 (\texttt{40-FIR}), and then approximates $\beta$ by the $\Hinf$-norm of the model.
To average the squared error, we use 10 Monte Carlo simulations, each of them running $10^5$ rounds.
The results are illustrated in Fig.~\ref{fig:double}b, where we observe that the proposed method (\WTS+ $\hat{\beta}_t$) attains the best performance among all algorithms.
We observe that, as predicted by~\cite{rojas-12}, \PI leads to asymptotically biased estimations under the presence of noise.
On the other hand, model-based approaches lead inherently to biased estimations since their model structure does not (in general) include the one generating the data.
The main source of confusion for the other three algorithms is the large variance at high frequency, which misleads these methods to believe that large outcomes are a consequence of large gains at high frequency.

%% file: figures/system_small.pdf_tex
\begingroup%
  \makeatletter%
  \providecommand\color[2][]{%
    \errmessage{(Inkscape) Color is used for the text in Inkscape, but the package 'color.sty' is not loaded}%
    \renewcommand\color[2][]{}%
  }%
  \providecommand\transparent[1]{%
    \errmessage{(Inkscape) Transparency is used (non-zero) for the text in Inkscape, but the package 'transparent.sty' is not loaded}%
    \renewcommand\transparent[1]{}%
  }%
  \providecommand\rotatebox[2]{#2}%
  \newcommand*\fsize{\dimexpr\f@size pt\relax}%
  \newcommand*\lineheight[1]{\fontsize{\fsize}{#1\fsize}\selectfont}%
  \ifx\svgwidth\undefined%
    \setlength{\unitlength}{180.6416513bp}%
    \ifx\svgscale\undefined%
      \relax%
    \else%
      \setlength{\unitlength}{\unitlength * \real{\svgscale}}%
    \fi%
  \else%
    \setlength{\unitlength}{\svgwidth}%
  \fi%
  \global\let\svgwidth\undefined%
  \global\let\svgscale\undefined%
  \makeatother%
  \begin{picture}(1,0.36174827)%
    \lineheight{1}%
    \setlength\tabcolsep{0pt}%
    \put(0,0){\includegraphics[width=\unitlength,page=1]{system_small.pdf}}%
    \put(0.27500183,0.04634099){\color[rgb]{0,0,0}\makebox(0,0)[lt]{\lineheight{0}\smash{\begin{tabular}[t]{l}$G$\end{tabular}}}}%
    \put(0.60395357,0.15898218){\color[rgb]{0,0,0}\makebox(0,0)[lt]{\lineheight{0}\smash{\begin{tabular}[t]{l}$e_\tau$\end{tabular}}}}%
    \put(0.81098493,0.05113798){\color[rgb]{0,0,0}\makebox(0,0)[lt]{\lineheight{0}\smash{\begin{tabular}[t]{l}$y_\tau$\end{tabular}}}}%
    \put(0.2628242,0.33796728){\color[rgb]{0,0,0}\makebox(0,0)[lt]{\lineheight{0}\smash{\begin{tabular}[t]{l} \end{tabular}}}}%
    \put(0,0){\includegraphics[width=\unitlength,page=2]{system_small.pdf}}%
    \put(-0.42361768,0.66204994){\color[rgb]{0,0,0}\makebox(0,0)[lt]{\begin{minipage}{1.46145689\unitlength}\raggedright  \end{minipage}}}%
    \put(-0.03093206,0.05228428){\color[rgb]{0,0,0}\makebox(0,0)[lt]{\lineheight{0}\smash{\begin{tabular}[t]{l}$u_\tau$\end{tabular}}}}%
    \put(0.5432532,0.04613179){\color[rgb]{0,0,0}\makebox(0,0)[lt]{\lineheight{0}\smash{\begin{tabular}[t]{l}$+$\end{tabular}}}}%
    \put(0,0){\includegraphics[width=\unitlength,page=3]{system_small.pdf}}%
    \put(0.23607489,0.282508){\color[rgb]{0,0,0}\makebox(0,0)[lt]{\lineheight{0}\smash{\begin{tabular}[t]{l}$w_\tau$\end{tabular}}}}%
    \put(0,0){\includegraphics[width=\unitlength,page=4]{system_small.pdf}}%
    \put(0.54230056,0.27561397){\color[rgb]{0,0,0}\makebox(0,0)[lt]{\lineheight{0}\smash{\begin{tabular}[t]{l}$H$\end{tabular}}}}%
  \end{picture}%
\endgroup%

%% file: figures/double3.pdf_tex
\begingroup%
  \makeatletter%
  \providecommand\color[2][]{%
    \errmessage{(Inkscape) Color is used for the text in Inkscape, but the package 'color.sty' is not loaded}%
    \renewcommand\color[2][]{}%
  }%
  \providecommand\transparent[1]{%
    \errmessage{(Inkscape) Transparency is used (non-zero) for the text in Inkscape, but the package 'transparent.sty' is not loaded}%
    \renewcommand\transparent[1]{}%
  }%
  \providecommand\rotatebox[2]{#2}%
  \newcommand*\fsize{\dimexpr\f@size pt\relax}%
  \newcommand*\lineheight[1]{\fontsize{\fsize}{#1\fsize}\selectfont}%
  \ifx\svgwidth\undefined%
    \setlength{\unitlength}{453.54330709bp}%
    \ifx\svgscale\undefined%
      \relax%
    \else%
      \setlength{\unitlength}{\unitlength * \real{\svgscale}}%
    \fi%
  \else%
    \setlength{\unitlength}{\svgwidth}%
  \fi%
  \global\let\svgwidth\undefined%
  \global\let\svgscale\undefined%
  \makeatother%
  \begin{picture}(1,0.35625)%
    \lineheight{1}%
    \setlength\tabcolsep{0pt}%
    \put(0.95292692,0.24171895){\color[rgb]{0,0,0}\makebox(0,0)[lt]{\begin{minipage}{0.33715019\unitlength}\raggedright  \end{minipage}}}%
    \put(0,0){\includegraphics[width=\unitlength,page=1]{double3.pdf}}%
    \put(0.10198506,0.34033013){\color[rgb]{0,0,0}\makebox(0,0)[lt]{\lineheight{0}\smash{\begin{tabular}[t]{l}\textbf{Frequency response}\end{tabular}}}}%
    \put(0.16414031,0.02315562){\color[rgb]{0,0,0}\makebox(0,0)[lt]{\lineheight{0}\smash{\begin{tabular}[t]{l}Arm $k$\end{tabular}}}}%
    \put(0,0){\includegraphics[width=\unitlength,page=2]{double3.pdf}}%
    \put(0.06358378,0.04733359){\color[rgb]{0.14901961,0.14901961,0.14901961}\makebox(0,0)[lt]{\lineheight{0}\smash{\begin{tabular}[t]{l}20\end{tabular}}}}%
    \put(0.1288746,0.04733359){\color[rgb]{0.14901961,0.14901961,0.14901961}\makebox(0,0)[lt]{\lineheight{0}\smash{\begin{tabular}[t]{l}60\end{tabular}}}}%
    \put(0.19151957,0.04733359){\color[rgb]{0.14901961,0.14901961,0.14901961}\makebox(0,0)[lt]{\lineheight{0}\smash{\begin{tabular}[t]{l}100\end{tabular}}}}%
    \put(0.25681037,0.04733359){\color[rgb]{0.14901961,0.14901961,0.14901961}\makebox(0,0)[lt]{\lineheight{0}\smash{\begin{tabular}[t]{l}140\end{tabular}}}}%
    \put(0.3221012,0.04733359){\color[rgb]{0.14901961,0.14901961,0.14901961}\makebox(0,0)[lt]{\lineheight{0}\smash{\begin{tabular}[t]{l}180\end{tabular}}}}%
    \put(0,0){\includegraphics[width=\unitlength,page=3]{double3.pdf}}%
    \put(0.01136748,0.06238212){\color[rgb]{0.14901961,0.14901961,0.14901961}\makebox(0,0)[lt]{\lineheight{0}\smash{\begin{tabular}[t]{l}0\end{tabular}}}}%
    \put(-0.00025537,0.17856415){\color[rgb]{0.14901961,0.14901961,0.14901961}\makebox(0,0)[lt]{\lineheight{0}\smash{\begin{tabular}[t]{l}0.5\end{tabular}}}}%
    \put(-0.00025537,0.28482428){\color[rgb]{0.14901961,0.14901961,0.14901961}\makebox(0,0)[lt]{\lineheight{0}\smash{\begin{tabular}[t]{l}1.0\end{tabular}}}}%
    \put(0,0){\includegraphics[width=\unitlength,page=4]{double3.pdf}}%
    \put(0.44065844,0.04788614){\color[rgb]{0.14901961,0.14901961,0.14901961}\makebox(0,0)[lt]{\lineheight{0}\smash{\begin{tabular}[t]{l}0\end{tabular}}}}%
    \put(0.5498118,0.04788614){\color[rgb]{0.14901961,0.14901961,0.14901961}\makebox(0,0)[lt]{\lineheight{0}\smash{\begin{tabular}[t]{l}2\end{tabular}}}}%
    \put(0.65896514,0.04788614){\color[rgb]{0.14901961,0.14901961,0.14901961}\makebox(0,0)[lt]{\lineheight{0}\smash{\begin{tabular}[t]{l}4\end{tabular}}}}%
    \put(0.76811852,0.04788614){\color[rgb]{0.14901961,0.14901961,0.14901961}\makebox(0,0)[lt]{\lineheight{0}\smash{\begin{tabular}[t]{l}6\end{tabular}}}}%
    \put(0.87727187,0.04788614){\color[rgb]{0.14901961,0.14901961,0.14901961}\makebox(0,0)[lt]{\lineheight{0}\smash{\begin{tabular}[t]{l}8\end{tabular}}}}%
    \put(0.96682221,0.04788614){\color[rgb]{0.14901961,0.14901961,0.14901961}\makebox(0,0)[lt]{\lineheight{0}\smash{\begin{tabular}[t]{l}10\end{tabular}}}}%
    \put(0.95606504,0.02981112){\color[rgb]{0.14901961,0.14901961,0.14901961}\makebox(0,0)[lt]{\lineheight{0}\smash{\begin{tabular}[t]{l}$\times 10^4$\end{tabular}}}}%
    \put(0,0){\includegraphics[width=\unitlength,page=5]{double3.pdf}}%
    \put(0.4006916,0.09951537){\color[rgb]{0.14901961,0.14901961,0.14901961}\makebox(0,0)[lt]{\lineheight{0}\smash{\begin{tabular}[t]{l}$10^{\text{-}6}$\end{tabular}}}}%
    \put(0.4006916,0.15584502){\color[rgb]{0.14901961,0.14901961,0.14901961}\makebox(0,0)[lt]{\lineheight{0}\smash{\begin{tabular}[t]{l}$10^{\text{-}4}$\end{tabular}}}}%
    \put(0.4006916,0.21217464){\color[rgb]{0.14901961,0.14901961,0.14901961}\makebox(0,0)[lt]{\lineheight{0}\smash{\begin{tabular}[t]{l}$10^{\text{-}2}$\end{tabular}}}}%
    \put(0.4006916,0.26850428){\color[rgb]{0.14901961,0.14901961,0.14901961}\makebox(0,0)[lt]{\lineheight{0}\smash{\begin{tabular}[t]{l}$10^0$\end{tabular}}}}%
    \put(0.40044817,0.32483393){\color[rgb]{0.14901961,0.14901961,0.14901961}\makebox(0,0)[lt]{\lineheight{0}\smash{\begin{tabular}[t]{l}$10^2$\end{tabular}}}}%
    \put(0.62527274,0.34148908){\color[rgb]{0,0,0}\makebox(0,0)[lt]{\lineheight{0}\smash{\begin{tabular}[t]{l}\textbf{Mean-squared error}\end{tabular}}}}%
    \put(0.62709804,0.0283371){\color[rgb]{0.14901961,0.14901961,0.14901961}\makebox(0,0)[lt]{\lineheight{0}\smash{\begin{tabular}[t]{l}Number of rounds $T$\end{tabular}}}}%
    \put(0.40896746,0.34056008){\color[rgb]{0,0,0}\makebox(0,0)[lt]{\begin{minipage}{0.00398197\unitlength}\raggedright  \end{minipage}}}%
    \put(1.48029665,0.21721851){\color[rgb]{0,0,0}\makebox(0,0)[lt]{\lineheight{0}\smash{\begin{tabular}[t]{l} \end{tabular}}}}%
    \put(0.17915429,0.00344495){\color[rgb]{0,0,0}\makebox(0,0)[lt]{\lineheight{0}\smash{\begin{tabular}[t]{l}(a)\end{tabular}}}}%
    \put(0.71824291,0.00344495){\color[rgb]{0,0,0}\makebox(0,0)[lt]{\lineheight{0}\smash{\begin{tabular}[t]{l}(b)\end{tabular}}}}%
    \put(0,0){\includegraphics[width=\unitlength,page=6]{double3.pdf}}%
    \put(0.27478522,0.22238333){\color[rgb]{0,0,0}\makebox(0,0)[lt]{\lineheight{0}\smash{\begin{tabular}[t]{l}$\norm{G}_\infty=0.7$\end{tabular}}}}%
    \put(0,0){\includegraphics[width=\unitlength,page=7]{double3.pdf}}%
    \put(0.8550047,0.28782034){\color[rgb]{0,0,0}\makebox(0,0)[lt]{\lineheight{0}\smash{\begin{tabular}[t]{l}\WTS: $\hat{\beta}_t$\end{tabular}}}}%
    \put(0,0){\includegraphics[width=\unitlength,page=8]{double3.pdf}}%
    \put(0.8550047,0.26145509){\color[rgb]{0,0,0}\makebox(0,0)[lt]{\lineheight{0}\smash{\begin{tabular}[t]{l}PI~\cite{rojas-12}\end{tabular}}}}%
    \put(0,0){\includegraphics[width=\unitlength,page=9]{double3.pdf}}%
    \put(0.8550047,0.23338965){\color[rgb]{0,0,0}\makebox(0,0)[lt]{\lineheight{0}\smash{\begin{tabular}[t]{l}10-FIR~\cite{tubore18}\end{tabular}}}}%
    \put(0,0){\includegraphics[width=\unitlength,page=10]{double3.pdf}}%
    \put(0.85624212,0.20532417){\color[rgb]{0,0,0}\makebox(0,0)[lt]{\lineheight{0}\smash{\begin{tabular}[t]{l}40-FIR~\cite{tubore18}\end{tabular}}}}%
    \put(0,0){\includegraphics[width=\unitlength,page=11]{double3.pdf}}%
    \put(0.08414472,0.30754908){\color[rgb]{0,0,0}\makebox(0,0)[lt]{\lineheight{0}\smash{\begin{tabular}[t]{l}$\abs{G(\ejwk)}$\end{tabular}}}}%
    \put(0,0){\includegraphics[width=\unitlength,page=12]{double3.pdf}}%
    \put(0.08414472,0.26910223){\color[rgb]{0,0,0}\makebox(0,0)[lt]{\lineheight{0}\smash{\begin{tabular}[t]{l}$\abs{H(\ejwk)}$\end{tabular}}}}%
    \put(0,0){\includegraphics[width=\unitlength,page=13]{double3.pdf}}%
  \end{picture}%
\endgroup%

%% file: contents/6_Conclusions.tex
\label{sec:6_Conclusions}

We have introduced a new instance of the stochastic multi-armed bandit problem in which the agent can play many arms per round by spreading a budget usually allocated to only one arm.
This setup is called a bandit problem under weighted information.
By spreading the available resources, the agent receives a Gaussian sequence as outcome, where the variance of each entry is inversely-proportional to the amount of resources allocated to that arm.
The goal of the agent is to minimize the regret of not allocating all the available resources to the arm that has the largest mean in magnitude.
We have derived a tight lower bound for the regret incurred by a wide class of bandit algorithms and introduced \WTS (Weighted Thompson Sampling), a Thompson-Sampling-based strategy whose regret upper bound has been shown, in this work, to match the predicted lower bound.
We have illustrated the importance of this new bandit problem by means of a fundamental problem in automatic control and system identification, also known as the \emph{gain estimation problem}, where the goal is to find the induced $\ell_2$-gain (or $\Hinf$-norm) of an unknown system in a model-free fashion.
By arguing that such a problem can be decomposed into a data-collection and point-estimation problems, we have shown that collecting data via \WTS and then using a point-estimator on this data outperforms state-of-the-art methods for solving this problem.
Future work involves studying the effect of weighted information from non-Gaussian distributions, as well as deriving problem-independent and probability bounds for the regret.

%% file: contents/Appendix/thm_lower.tex
Before proving this result, we need to introduce the following technical lemma~\cite{muller17}, which we re-state here for completeness, together with its respective proof.

\begin{lemma}
\label{lem:D>d}
Let $\nu$ and $\nu'$ denote two different distributions for a random variable, and let $\P$ and $\P'$ denote the respective probability measures.
Then, for every measurable event $B$, the \emph{KL}-divergence between $\nu$ and $\nu'$ satisfies
\begin{align}
\KL{\nu}{\nu'} \geq d(\P\{B\},\P'\{B\}),
\end{align}
where $d(q,s) = q\log(q/s) + (1-q)\log([1-q]/[1-s])$ is the binary entropy function~\cite{covtho06}, corresponding to the \emph{KL}-divergence between two Bernoulli variables of means $q$ and $s$.
\end{lemma}

\begin{proof}
Let  $\mean{\cdot}$ and $\meanp{\cdot}$ denote the expectation operator under $\P$ and $\P'$, respectively.
Introducing a change of measure, and by Jensen's inequality~\cite{hardy-88}, we have that
\begin{align}
\P'\{B\} = \meanp{\one{B}} = \int\left(\frac{d\P'}{d\P}\right)\one{B}d\P 
&=
 \int \exp{\ell(\nu')-\ell(\nu)}\one{B}d\P
 \notag\\
&=
\mean{\exp{\ell(\nu')-\ell(\nu)}\one{B}}
\notag\\
&=\mean{\exp{\ell(\nu')-\ell(\nu)}\,\big|\, B}\P\{B\}
\notag\\
&\geq
\exp{\mean{\ell(\nu')-\ell(\nu)|B}}\P\{B\},
\end{align}
from which we conclude that
\begin{align}
\mean{\ell(\nu)-\ell(\nu')|B} \geq \log\frac{\P\{B\}}{\P'\{B\}}.
\end{align}
It now follows that
\begin{align}
\KL{\nu}{\nu'} &= \mean{\ell(\nu)-\ell(\nu')}
\notag\\
&= \mean{\ell(\nu)-\ell(\nu')|B}\P\{B\} + \mean{\ell(\nu)-\ell(\nu')|B^c}\P\{B^c\}
\notag\\
&\geq
\P\{B\}\log\frac{\P\{B\}}{\P'\{B\}} + \P\{B^c\}\log\frac{\P\{B^c\}}{\P'\{B^c\}}
\notag\\
&=
\P\{B\}\log\frac{\P\{B\}}{\P'\{B\}} + (1-\P\{B\})\log\frac{1-\P\{B\}}{1-\P'\{B\}}
\notag\\
&=
d(\P\{B\},\P'\{B\}).
\end{align}
\end{proof}

To prove the regret lower bound, we follow the standard approach by~\cite{lairob85,burkat96}.

Without loss of generality, let $\bmu\eq(\bmu_1,\dots,\bmu_K)$ satisfy $\norm{\bmu_1}>\norm{\bmu_k}$, $k=2,\dots,K$, meaning that arm 1 is assumed to be the (unique) optimal arm.
The log-likelihood function of $\bmu$ and $\bsigma$ given the outcomes $(X^1,X^2,\dots,X^T)$ is
\begin{align}
\ell(\bmu,\bsigma) &= \log \nu(X^1,\dots,X^t)
\notag\\
&=
\sum_{t=1}^T\sum_{k=1}^K \log\left(\frac{p^{t,\bpi}_k}{\pi\sigma_k^2}\exp{-\frac{p^{t,\bpi}_k}{\sigma_k^2}\norm{X^t_k-\bmu_k}^2} \right)
\notag\\
&=
\sum_{t=1}^T\sum_{k=1}^K
\left( -\log\pi + \log p^{t,\bpi}_k - \log\sigma^2_k - \frac{p^{t,\bpi}_k\norm{X^t_k-\bmu_k}^2}{\sigma^2_k}\right).
\end{align}
Now consider confusion parameters $\bmu'\eq(\bmu_1',\dots,\bmu_K')$ and $\bsigma'\eq(\sigma_1'^2,\dots,\sigma_K'^2)$, and let $\Psp_{t+1}$ denote the $\sigma$-algebra generated by $(p^1,p^2,\dots,p^t)$.
Then, the expected difference of the likelihood function at $(\bmu,\bsigma)$ and at $(\bmu',\bsigma')$ satisfies
\begin{align}
\mathbb{E}_{\bmu,\bsigma}\big\{
\ell(\bmu,&\bsigma)-\ell(\bmu',\bsigma') \given\Psp_{t+1}\big\}
\notag\\
&=
\mathbb{E}_{\bmu,\bsigma}\left\{
\sum_{t=1}^T\sum_{k=1}^K
\left(
\frac{p^{t,\bpi}_k\norm{X^t_k-\bmu'_k}^2}{\sigma_k'^2} - \frac{p^{t,\bpi}_k\norm{X^t_k-\bmu_k}^2}{\sigma_k^2 } + \log\frac{\sigma_k'^2}{\sigma_k^2}
\right)
\Given\Psp_{t+1}
\right\}
\notag\\
&=
\sum_{t=1}^T\sum_{k=1}^K
\left(
\frac{p^{t,\bpi}_k \left( \frac{\sigma^2_k}{p^{t,\bpi}_k} + \norm{\bmu_k-\bmu_k'}^2\right) }{\sigma_k'^2} - 1 + \log\frac{\sigma_k'^2}{\sigma_k^2}
\right)
\notag\\
&=
\sum_{t=1}^T\sum_{k=1}^K
\left(
\frac{p^{t,\bpi}_k\norm{\bmu_k-\bmu_k'}^2}{\sigma_k'^2} - 1 +\frac{\sigma_k^2}{\sigma_k'^2}  +\log\frac{\sigma_k'^2}{\sigma_k^2}\right),
\end{align}
where $\mathbb{E}_{\bmu,\bsigma}$ denotes the expectation under the distribution of the outcomes determined by $(\bmu,\bsigma)$.
Now, let parameter $\bmu'$ satisfy $\bmu_k'=\bmu_k$, for every $k\neq a$, and $\norm{\bmu_a'} >\norm{\bmu_1}$, for a fixed $a\neq 1$.
This means that arm 1 is optimal in $(\bmu,\bsigma)$ but suboptimal in $(\bmu',\bsigma')$.
It then follows that
\begin{align}
\KL{\bmu,\bsigma}{\bmu',\bsigma'} 
&=
\mean{\ell(\bmu,\bsigma)-\ell(\bmu,\bsigma')}
\notag\\
&=
\mean{\mathbb{E}_{\bmu,\bsigma}\left\{
\ell(\bmu,\bsigma)-\ell(\bmu,\bsigma_k')
\given \Psp_{t+1} \right\}}
\notag\\
&=
\frac{\mean{p^{t,\bpi}_a}\norm{\bmu_a-\bmu_a'}^2}{\sigma_a'^2} - 1 +\frac{\sigma_a^2}{\sigma_a'^2}  + \log\frac{\sigma_a'^2}{\sigma_a^2}.
\label{eq:kl}
\end{align}
On the other hand, let $\P$ and $\P'$ (and $\mean{\cdot}$ with $\meanp{\cdot}$) denote probability measures for $(X^t)_t$ (and expectations) under $(\bmu,\bsigma)$ and $(\bmu',\bsigma')$, respectively, and recall that $\bpi\in\bPi^\star$.
Invoking Lemma~\ref{lem:D>d} with $B = B_T \eq \{\sum_{t=1}^T p_1^{t,\bpi}<T-T^\gamma\}$, for some $0<\gamma<1$, yields
\begin{align}
\P\{B_T\} = \P\left\{T-{\textstyle\sum_{t=1}^T} p_1^{t,\bpi}>T^\gamma\right\}
&\overset{(a)}{=} \P\left\{ {\textstyle\sum_{k=2}^K\sum_{t=1}^T} p^{t,\bpi}_k > T^\gamma\right\}
\notag\\
&\overset{(b)}{\leq} \frac{1}{T^\gamma} {\textstyle\sum_{k=2}^K\sum_{t=1}^T} \mean{p^{t,\bpi}_k}
\notag\\
&\overset{(c)}{=} \o(T^{\alpha-\gamma}),
\label{eq:P}
\end{align}
for every $\alpha > 0$, where $(a)$ follows from $\sum_{k=1}^K p^{t,\bpi}_k = 1$ for every $t\in\setT$, $(b)$ is Markov's inequality~\cite{durret10}, and where $(c)$ is a consequence of $\bpi$ being uniformly efficient according to Definition~\ref{def:uniform}, \ie, the expected-cumulative power allocated at arms $2,\dots,K$ by $\bpi\in\bPi^\star$ is $\o(T^\alpha)$.
Similarly, under $(\bmu,\bsigma')$ (where arm $1$ is suboptimal) we have that
\begin{align}
\P'\{B^c_T\} = \P'\left\{ T-{\textstyle\sum_{t=1}^T} p_1^{t,\bpi} < T^\gamma\right\}
&\leq \frac{1}{T-T^\gamma} {\textstyle \sum_{t=1}^T}\meanp{p^{t,\bpi}_1}
\notag\\
&= \frac{\o(T^\alpha)}{T-T^{\alpha-1}}
\notag\\
&= \o(T^{\alpha-1}),
\label{eq:P'}
\end{align}
for every $\alpha > 0$.
Now for $\alpha <\gamma$,~\eqref{eq:P} and~\eqref{eq:P'} lead us to
\begin{align}
d(\P\{B_T\},\P'\{B_T\})
&\geq
\P\{B_T\}\log\frac{\P\{B_T\}}{\P'\{B_T\}} + (1-\P\{B_T\})\log\frac{1-\P\{B_T\}}{1-\P'\{B_T\}}
\notag\\
&=
\o(T^{\alpha-\gamma})\log\frac{\o(T^{\alpha-\gamma})}{1-\o(T^{\alpha-1})}
+ (1-\o(T^{\alpha-\gamma}))\log\frac{1- \o(T^{\alpha-\gamma})}{\o(T^{\alpha-1})}
\notag\\
&\geq
\o(1) + (1-\alpha)\log T.
\end{align}
Thus, combining this result with~\eqref{eq:kl} and Lemma~\ref{lem:D>d} yields
\begin{align}
\sum_{t=1}^T
\frac{\mean{p^{t,\bpi}_a}\norm{\bmu_a-\bmu_a'}^2}{\sigma_a'^2} - 1 +\frac{\sigma_a^2}{\sigma_a'^2}  + \log\frac{\sigma_a'^2}{\sigma_a^2}
&\geq \o(1) + (1-\alpha)\log T,
\end{align}
implying hat
\begin{align}
\sum_{t=1}^T\mean{p^{t,\bpi}_a} 
&\geq 
\frac{\sigma_a'^2}{\norm{\bmu_a-\bmu_a'}^2}\left[
\left(1 -\frac{\sigma_a^2}{\sigma_a'^2}  -\log\frac{\sigma_a'^2}{\sigma_a^2}\right)T + \o(1) + (1-\alpha)\log T
\right].
\label{eq:lb}
\end{align}
It now follows that the confusion parameters $(\bmu',\bsigma')$ maximizing the lower bound in~\eqref{eq:lb} are obtained by solving
\begin{align}
\sup_{\bmu_a',\sigma_a'^2\colon\norm{\bmu'_a}>\norm{\bmu_1}}
\frac{\sigma_a'^2}{\norm{\bmu_a-\bmu_a'}^2}\left[
\left(1 -\frac{\sigma_a^2}{\sigma_a'^2}  -\log\frac{\sigma_a'^2}{\sigma_a^2}\right)T + \o(1) + (1-\alpha)\log T
\right],
\notag
\end{align}
where a necessary condition for optimality is that the term multiplying $T$ must be zero since 
$$1-\frac{\sigma_a^2}{\sigma_a'^2} - \log\frac{\sigma_a'^2}{\sigma_a^2} \leq 0,$$
which is achieved iff $\sigma^2_a=\sigma_a'^2$.
The supremum is attained by choosing $\bmu'_a = \norm{\bmu_1} \bmu_a/\norm{\bmu_a}$, that is, a vector in the direction of $\bmu_a$ with the magnitude of $\bmu_1$, allowing us to conclude
\begin{align}
\liminf_{T\to\infty}\frac{\sum_{t=1}^T \mean{p^{t,\bpi}_a}}{\log T} \geq \frac{\sigma_a^2(1-\alpha)}{(\norm{\bmu_1}-\norm{\bmu_a})^2} = \frac{\sigma_a^2(1-\alpha)}{\Delta_a^2}.
\end{align}
The above result is valid for every $a=2,3,\dots,K$, so the proof is finished by taking $\alpha\to 0^+$ and observing that $\mean{\regret^{\bpi}_{\bmu,\bsigma}(T)} = \sum_{a=2}^K \sum_{t=1}^T\mean{p^{t,\bpi}_a} \Delta_a$.

%% file: contents/Appendix/lem_suff.tex
To prove this result, we first show that $\xx_k(t)$, $S_k(t)$ and $(p^1,p^2,\dots,p^t)$ are sufficient statistics for the distribution of the outcomes.
This is followed by the characterization of their distributions and statistical independence.

\subsection{Sufficiency of the statistics}

The density of the sequence $(X^t_k)_t$, for every arm $k\in\setK$, is given by
\begin{align}
&\nu(X^1_k,\dots,X^t_k\given p^1_k,\dots,p^t_k) 
\notag\\
&\hspace{2.4cm}= 
\frac{(\prod_{\ell=1}^t p^\ell_k)}{(\pi\sigma_k^2)^n}\exp{-\frac{1}{\sigma_k^2}\sum_{\ell=1}^t p^\ell_k\norm{X^\ell_k-\bmu_k}^2}
\notag\\
&\hspace{2.4cm}=
\frac{(\prod_{\ell=1}^t p^\ell_k)}{(\pi\sigma_k^2)^n}\exp{-\frac{1}{\sigma_k^2}\sum_{\ell=1}^t p^\ell_k\norm{X^\ell_k-\bmu_k + \xx_k(t)-\xx_k(t)}^2}
\notag\\
&\hspace{2.4cm}=
\frac{(\prod_{\ell=1}^t p^\ell_k)}{(\pi\sigma_k^2)^n}\exp{-\frac{1}{\sigma_k^2}\sum_{\ell=1}^t p^\ell_k
(\norm{X^\ell_k-\xx_k(t)}^2 + \norm{\xx_k(t)-\bmu_k}^2 -2 [X^\ell_k-\xx_k(t)]^\top[\bmu_k -\xx_k(t)])},
\notag
\end{align}
where
\begin{align}
&\sum_{\ell=1}^t p^\ell_k [X^\ell_k-\xx_k(t)]^\top[\bmu_k -\xx_k(t)]
\notag\\
&\hspace{1.2cm}\hspace{1.3cm}=
\sum_{\ell=1}^t \left(X^\ell_k-\frac{\sum_{i=1}^t p^i_k X^i_k }{\sum_{j=1}^t p^j_k}\right)^\top\left(\bmu_k  -\frac{\sum_{i=1}^t p^i_k X^i_k }{\sum_{j=1}^t p^j_k}\right)
\notag\\
&\hspace{1.2cm}\hspace{1.3cm}=
\frac{1}{\left(\sum_{j=1}^t p^j_t\right)^2}
\sum_{\ell=1}^t p^\ell_k
\left(\sum_{j=1}^t p^j_k X^\ell_k-\sum_{i=1}^t p^i_k X^i_k \right)^\top\left(\sum_{j=1}^t p^j_k\bmu_k  -\sum_{i=1}^t p^i_k X^i_k \right)
\notag\\
&\hspace{1.2cm}\hspace{1.3cm}=
\frac{1}{\left(\sum_{j=1}^t p^j_t\right)^2}
\sum_{\ell=1}^t \sum_{i=1}^t \sum_{j=1}^t
p^\ell_k p^i_k p^j_k (X^\ell_k - X^i_k)^\top(\bmu_k-X^j_k)
\notag\\
&\hspace{1.2cm}\hspace{1.3cm}=
\frac{1}{\left(\sum_{j=1}^t p^j_t\right)^2}
\sum_{j=1}^t p^j_k(\bmu_k-X^j_k)^\top \underbrace{\sum_{\ell=1}^t \sum_{i=1}^t p^\ell_k p^i_k  (X^\ell_k - X^i_k)}_{=0}
\notag\\
&\hspace{1.2cm}\hspace{1.3cm}=
0.
\end{align}
Therefore,
\begin{align}
\nu(X^1_k,\dots,X^t_k\given p^1_k,\dots,p^t_k) 
&=
\frac{(\prod_{\ell=1}^t p^\ell_k)}{(\pi\sigma_k^2)^n}\exp{-\frac{1}{\sigma_k^2}\sum_{\ell=1}^t p^\ell_k
(\norm{X^\ell_k-\xx_k(t)}^2 + \norm{\xx_k(t)-\bmu_k}^2)}
\notag\\
&=
\frac{(\prod_{\ell=1}^t p^\ell_k)}{(\pi\sigma_k^2)^n}\exp{-\frac{1}{\sigma_k^2}\left(S_k(t) + \norm{\xx_k(t)-\bmu_k}^2\sum_{\ell=1}^t p^\ell_k\right)},
\end{align}
which means that it is possible to keep track of the distribution of the sequence $(X^1_k,\dots,X^t_k)$ (given the trajectory of allocated power) by just considering $S_k(t)$ and $\xx_k(t)$, for every $k\in\setK$.

\subsection{Distributions and statistical independence}

To simplify the notation, we write $p^t_k$ instead of $p^{t,\WTS}_k$.
Additionally, in this section, all the expectations (and covariance matrices) are conditioned on $\Psp_{t+1}=\sigma(p^1_k,p^2_k,\dots,p^t_k)$.
Notice that $\xx_k(t)$ is Gaussian distributed, since $(X^1_k,\dots,X^t_k)$ is an independent Gaussian sequence, with mean $\mean{\xx_k(t)} = \bmu_k$ and whose covariance matrix is
\begin{align}
\cov\{\xx_k(t)\}
&=
\mean{(\xx_k(t)-\mu_k)(\xx_k(t)-\mu_k)^\top}
\notag\\
&=
\frac{1}{(\sump)^2}\mean{\left(\sum_{i=1}^t p^i_k(X^i_k-\bmu_k)\right)\left(\sum_{i=1}^t p^i_t(X^i_k-\bmu_k)\right)^\top}
\notag\\
&=
\frac{1}{(\sump)^2}\mean{\sum_{i=1}^t (p^i_k)^2(X^i_k-\bmu_k)(X^i_k-\bmu_k)^\top}
\notag\\
&=
\frac{1}{(\sump)^2}\sum_{i=1}^t (p^i_k)^2\frac{\sigma_k^2}{2p^i_k}\bI_2
\notag\\
&=
\frac{\sigma^2_k}{2\sump}\bI_2.
\end{align}

To characterize the distribution of $S_k(t) = \sum_{\ell=1}^t p^\ell_k \norm{X^\ell_k-\xx_k(t)}^2$ we exploit the fact that the data is statistically independent across arms and rounds.
To this end, let us rewrite $S_k(t) = \sum_{\ell=1}^t \left[p^\ell_k (X_t^k(1)-\xx_k(t,1))^2 + p^\ell_k(X_t^k(2)-\xx_k(t,2))^2\right]$, where $X^t_k(i)$ and $\xx_k(t,i)$, $i\in\{1,2\}$ denote the $i$-th component of the $\R^2$-vectors $X^t_k$ and $\xx_k(t)$, respectively.
Since the terms $(X^t_k(i)-\xx_k(t,i))^2$ in $S_k(t)$ are iid, we only need to characterize one of them.
Let us introduce $s_k(t) \eq \sum_{\ell=1}^t p^\ell_k(X^\ell_k(1)-\xx_k(t,1))^2$ and, for $k$ and $t$ fixed, let $\bX\eq[X^1_k(1)\quad\cdots\quad X^t_k(1)]^\top$, $\bp\eq[p^1_k\quad\cdots\quad p^t_k]^\top$.
Additionally, let $\bmu_k(1)$ denote the first component of $\bmu_k\in\R^2$.
Notice that $(\bX-\bmu_k(1)\bone)\sim\Nsp(\bzero_t, (\sigma_k^2/2)\diag\bp)$, for every $i=1,\dots,t$, 
since the outcomes $(X^1_k,X^2_k,\dots,X^t_k)$ are Gaussian, statistically independent, and with individual distribution $X^t_k(1)\sim\Nsp(\bmu_k(1),\sigma_k^2/2)$.
Under this definition, $\xx_k(t,1) = \bone^\top\bX/(\bone\bp)$ and, hence, the first component of $\bmu_k\in\R^2$, $s_k(t)$ can be written as
\begin{align}
s_k(t) 
&=
\bX^\top \left(\bI_t - \frac{\bone\bp^\top}{\bone^\top\bp} \right)^\top\diag{\bp}\left(\bI_t - \frac{\bone\bp^\top}{\bone^\top\bp} \right)\bX
\notag\\
&=
\bX^\top\left( \diag\bp -\frac{\bp\bp^\top}{\bone^\top\bp}\right)\bX
\notag\\
&=
\frac{\sigma_k^2}{2}\bE^\top\left(\diag\bp\right)^{-1/2} \left( \diag\bp -\frac{\bp\bp^\top}{\bone^\top\bp}\right)\left(\diag\bp\right)^{-1/2}\bE
\notag\\
&=\frac{\sigma_k^2}{2}\bE^\top\underbrace{\left(\bI_t-\frac{\bp^{1/2}\bp^{\top/2}}{\bone^\top\bp}\right)}_{\eq\bA}\bE,
\end{align}
where $\bE\sim\Nsp(\bzero_t,\bI_t)$, and where $\bp^{1/2}\eq[\hsqrt{p^1_k}\quad \dots \quad \hsqrt{p^t_k}]^\top$.
Notice that $\textnormal{Rank}\,\bA=t-1$ since $\bp^{1/2}\bp^{\top/2}$ is a rank-1 perturbation~\cite{horjoh12}, and that $\bA$ corresponds to an idempotent matrix since
\begin{align}
\left(\bI_t-\frac{\bp^{1/2}\bp^{\top/2}}{\bone^\top\bp}\right)^2 = \bI_t-2\frac{\bp^{1/2}\bp^{\top/2}}{\bone^\top\bp}+\frac{\bp^{1/2}\bp^{\top/2}\bp^{1/2}\bp^{\top/2}}{(\bone^\top\bp)^2}
=
\bI_t-\frac{\bp^{1/2}\bp^{\top/2}}{\bone^\top\bp}.
\end{align}
Then, by~\cite[Lemma B.2]{soders02}, we have that $s_k(t)/(\sigma_k^2/2)\sim\chi^2_{t-1}$ and therefore, because both terms in $S_k(t)$ are iid, $S_k(t)/(\sigma_k^2/2)\sim\chi^2_{2(t-1)}$.

It now remains to prove conditional independence of $\xx_k(t)$ and $S_k(t)$, for every $k\in\setK$, and $t\in\setT$, given the trajectory of the power profiles.
To show this, observe that
\begin{align}
\cov{(X^i_k-\xx_k(t),\xx_k(t)})
&=
\mean{(X^i_k-\xx_k(t))(\xx_k(t)-\bmu_k)^\top}
\notag\\
&=(\sump)^{-2}\mean{\sum_{j=1}^t\sum_{\ell=1}^t p_j p_\ell (X^i_k-X^j_k)(X_\ell-\bmu_k)}
\notag\\
&=
0,
\end{align}
and that $X^i_k$ and $\xx_k(t)$ are jointly Gaussian distributed.
This implies that $\{X^i_k-\xx_k(t)\}_{i=1}^t$ are statistically independent of $\xx_k(t)$ and, therefore, the result follows because $S_k(t)$ is a function of the former.

%% file: contents/Appendix/lem_post.tex
Recall that, from Lemma~\ref{lem:WTS_suff}, $\xx_k(t)$ and $S_k(t)$ are conditionally independent (given the trajectory of allocated powers).
In consequence, for every $k\in\setK$ and $t=3,4,\dots$, the posterior distribution of $(\bmu_k,\sigma_k^2)$ given $(\xx_k(t-1),{S_k(t-1)},p^1_k,p^2_k,\dots,p^{t-1}_k)$ is 
\begin{align}
&f_{\bmu_k,\sigma_k^2\given\xx_k(t-1)=\bx,S_k(t-1)=s,p^1_k,p^2_k,\dots,p^{t-1}_k}(\bmu_k,\sigma_k^2)
\notag\\
&\hspace{1cm}=
\frac{f_{\xx_k(t-1),S_k(t-1)\given\bmu_k,\sigma_k^2,p^1_k,p^2_k,\dots,p^{t-1}_k}(\bx,s)\cdot f_{\bmu_k,\sigma_k\given p^1_k,p^2_k,\dots,p^{t-1}_k} (\bmu_k,\sigma_k^2) }
{\int_0^\infty \!\!\! \int_{\R^2} f_{\xx_k(t-1),S_k(t-1)\given\bmu_k,\sigma_k^2,p^1_k,p^2_k,\dots,p^{t-1}_k}(\bx,s)\cdot f_{\bmu_k,\sigma_k\given p^1_k,p^2_k,\dots,p^{t-1}_k} (\bmu_k,\sigma_k^2) d\bmu\,d(\sigma^2)}
\notag\\
&\hspace{1cm}=
\frac{f_{\xx_k(t-1),S_k(t-1)\given\bmu_k,\sigma_k^2,p^1_k,p^2_k,\dots,p^{t-1}_k}(\bx,s)\cdot \overbrace{f_{\bmu_k,\sigma_k\given \Fsp_1} (\bmu_k,\sigma_k^2)}^{\propto 1} }
{\int_0^\infty \!\!\!  \int_{\R^2} f_{\xx_k(t-1),S_k(t-1)\given\bmu_k,\sigma_k^2,p^1_k,p^2_k,\dots,p^{t-1}_k}(\bx,s)\cdot f_{\bmu_k,\sigma_k\given \Fsp_1} (\bmu_k,\sigma_k^2) d\bmu\,d(\sigma^2)}
\notag\\
&\hspace{1cm}=
\frac{f_{\xx_k(t-1),S_k(t-1)\given\bmu_k,\sigma_k^2,p^1_k,p^2_k,\dots,p^{t-1}_k}(\bx,s) }
{\int_0^\infty \!\!\! \int_{\R^2} f_{\xx_k(t-1),S_k(t-1)\given\bmu_k,\sigma_k^2,p^1_k,p^2_k,\dots,p^{t-1}_k}(\bx,s) d\bmu\,d(\sigma^2)}
\label{eq:int}
\end{align}
where, from Lemma~\ref{lem:WTS_suff},
\begin{align}
f_{\xx_k(t-1),S_k(t-1)\given\bmu_k,\sigma_k^2,p^1_k,p^2_k,\dots,p^{t-1}_k}(\bx,s)
=
\frac{(\sump)s^{t-2}}{\pi\Gamma(t-1)}\frac{\exp{\frac{-1}{\sigma_k^2} \left(s+\sump\norm{\bx-\bmu_k}^2\right)} }{(\sigma_k^2)^t}.
\end{align}
Thus, the integral term in~\eqref{eq:int} is given by
\begin{align}
\int_0^\infty\int_{\R^2} &f_{\xx_k(t),S_k(t)\given\bmu_k=\bmu,\sigma_k^2=\sigma^2,p^1_k,p^2_k,\dots,p^{t-1}_k}(\bx,s)d\bmu\,d(\sigma^2)
\notag\\
&\hspace{3.5cm}=
\frac{(\sump)s^{t-2}}{\pi\Gamma(t-1)} \int_0^\infty \frac{\exp{-s/\sigma^2}}{(\sigma^2)^t}\int_{\R^2}\exp{\frac{-1}{\sigma^2}\sump\norm{\bx-\bmu}^2} d\bmu d(\sigma^2)
\notag\\
&\hspace{3.5cm}=
\frac{s^{t-2}}{\Gamma(t-1)}\int_0^\infty \frac{\exp{-s/\sigma^2}}{(\sigma^2)^{t-1}}d(\sigma^2)
\notag\\
&\hspace{3.5cm}=
\frac{\Gamma(t-2)}{\Gamma(t-1)},
\end{align}
implying that
\begin{align}
&f_{\bmu_k,\sigma_k^2\given\xx_k(t-1)=\bx,S_k(t-1)=s,p^1_k,p^2_k,\dots,p^{t-1}_k}(\bmu_k,\sigma_k^2)
=
\frac{\sump s^{t-2}}{\pi\Gamma(t-2)}\frac{\exp{\frac{-1}{\sigma_k^2}\left(s+\sump\norm{\bx-\bmu_k}^2\right)}}{(\sigma_k^2)^t}.
\notag
\end{align}
Therefore, the posterior distribution of $\bmu_k$ given $(\xx_k(t),S_k(t),p^1_k,p^2_k,\dots,p^{t-1}_k)$ is
\begin{align}
f_{\bmu_k\given\xx_k(t-1)=\bx,S_k(t-1)=s,p^1_k,p^2_k,\dots,p^{t-1}_k}(\bmu_k)
&=
\frac{\sump s^{t-2}}{\pi\Gamma(t-2)}\int_0^\infty \frac{\exp{\frac{-1}{\sigma^2}\left(s+\sump\norm{\bx-\bmu_k}^2\right)}}{(\sigma^2)^t} d(\sigma^2)
\notag\\
&=
\frac{\sump (t-2)}{\pi s}\left(1+\frac{\sump\norm{\bx-\bmu_k}^2}{s}\right)^{-t+1}.
\notag
\end{align}

%% file: contents/Appendix/thm_upper.tex
We proceed by decomposing the regret $R^{\WTS}_{\bmu,\bsigma}(T)$ into three different terms which we upper bound separately in three different lemmas presented in Section~\ref{app:three_lemmas}.
It is worth to mention that splitting the regret into these 3 terms is a standard technique in finite-time regret analysis, and our choice is inspired directly by~\cite{hontak14,mulroj19}.
Nevertheless, upper-bounding each of these terms involves completely different techniques than the ones employed in~\cite{hontak14,mulroj19} since the direct application of the techniques in those references to bandits under weighted information does not hold.

The technical results supporting the three upper bounds, such as concentration inequalities, can be found in Section~\ref{app:technical_lemmas}.

Without loss of generality, let $k^\star = 1$. 
Also, let $0<\e<\min_k(\norm{\bmu_1}-\norm{\bmu_k})/2$ and define the following events:
\begin{align}
\Asp_t &\eq \{\norm{\tbmu^\star(t)} \geq \norm{\bmu_1} - \e\},
\notag\\
\Bsp_{k,t} &\eq \{ \norm{\xx_k(t)-\bmu_k} \leq \e/2 \quad\text{and}\quad S_k(t) \leq t(\sigma^2_k + \e) \},
\notag
\end{align}
where $\tbmu_k(t)\sim f_{\bmu_k\given\Fsp_t}$ as in Lemma~\ref{lem:post_WTS}, and where $\norm{\tbmu^\star(t)} \eq \max_k \norm{\tbmu_k(t)}$.
Observe that the events $\Bsp_{k,t}$ are defined in terms of $\xx_k(t)$ and $S_k(t)$, unlike the ones in the proof of optimality of TS~\cite{mulroj19} where their counterparts $\Bsp_t(k)$ are defined in terms of $\xx_k{(t-1)}$ and $S_k(t-1)$.
This choice of events $\Bsp_{k,t}$ is essential in proving this theorem.

Since all prior mean distributions are equal, we have that $\rho^1_k=\rho^2_k = K^{-1}$, for every $k\in\setK$, according to Algorithm~\ref{alg:wts}.
By the definition of $p^{t,\WTS}_k$, it holds that
\begin{align}
\mean{p^{t,\WTS}_k}
=
\mean{\rho^t_k}
=
\mean{\prob{\tkstar_t = k\given\Fsp_t}}
=
\prob{\tkstar_t = k},
\end{align}
where $\prob{\tkstar_t = k}$ is the unconditional probability of arm $k$ being optimal.
Then, by~\eqref{eq:rho_wts}, the expected cumulative regret under WTS satisfies
\begin{align}
\mean{\regret^{\WTS}_{\bmu,\bsigma}(T)} 
&=
\mean{\sum_{t=1}^T \sum_{k=2}^K \Delta_k p^{t,\WTS}}
\notag\\
&=
\sum_{k=2}^K\Delta_k(\rho^1_k+\rho^2_k) + \mean{\sum_{t=3}^T \sum_{k=2}^K \Delta_k p^{t,\WTS}} 
\notag\\
&=
2K^{-1}\sum_{k=2}^K \Delta_k + \sum_{t=3}^T \sum_{k=2}^K \Delta_k \prob{\tkstar_t = k} 
\notag\\
&\leq
2\Delta_{\max} + \sum_{t=3}^T \sum_{k=2}^K \Delta_k \prob{\tkstar_t = k} 
\notag\\
&=
2\Delta_{\max} + \sum_{k=2}^K \Delta_k \left( \sum_{t=3}^T \prob{\tkstar_t = k,\Asp_t} + \prob{\tkstar_t = k,\Asp_t^c} \right)
\notag\\
&\leq
2\Delta_{\max} + \sum_{k=2}^K \Delta_k \Bigg(  \sum_{t=3}^T \prob{\tkstar_t = k,\Asp_t,\Bsp_{k,t}} 
+ \prob{\tkstar_t = k,\Bsp_{k,t}^c} 
\notag\\
&\hspace{8.5cm} + \prob{\tkstar_t = k,\Asp_t^c} \Bigg)
\notag\\
&\leq
2\Delta_{\max} + \sum_{k=2}^K \!\Delta_k  \sum_{t=3}^T \!\left(\prob{\tkstar_t = k,\Asp_t,\Bsp_{k,t}} \!+\! \prob{\tkstar_t = k,\Bsp_{k,t}^c} \right)
\notag\\
&\hspace{7.5cm} +  \Delta_{\max}\sum_{t=3}^T \prob{\tkstar_t \neq 1,\Asp_t^c} ,
\label{eq:proof_wts}
\end{align}
where $\Asp^c_t$ and $\Bsp_{k,t}^c$ denote the complements of $\Asp_t$ and $\Bsp_{k,t}$, respectively.
Now, by means of Lemma~\ref{lem:wts_first}, Lemma~\ref{lem:wts_second} and Lemma~\ref{lem:wts_third} (defined in the following section), the expected cumulative regret under WTS satisfies
\begin{align}
\mean{\regret^{\WTS}_{\bmu,\bsigma}(T) }
&\leq 
 \sum_{k=2}^K \Delta_k \frac{\sigma_k^2+\e}{(\Delta_k-2\e)^2}\log T + \O(1) + \O(\e^{-2}) + \O(\e^{-2}),
\end{align}
where the $\O$ terms are independent of $T$, implying that
\begin{align}
\frac{\mean{\regret^{\WTS}_{\bmu,\bsigma}(T)}}{\log T} \leq 
 \sum_{k=2}^K \Delta_k\frac{\sigma_k^2+\e}{(\Delta_k-2\e)^2}
+
\frac{\O(1) + \O(\e^{-2}) + \O(\e^{-2})}{\log T}.
\label{eq:many_e}
\end{align}
The numerator of the second term is independent of $T$, so the result now follows from choosing $\e = \e(T) < \log^{-a} T$, for some $0<a<1/2$ and taking the limit $T\to\infty$:
\begin{align}
\limsup_{T\to\infty}\frac{\mean{\regret^{\WTS}_{\bmu,\bsigma}(T)}}{\log T} \leq 
\sum_{k=2}^K \frac{\sigma_k^2}{\Delta_k}.
\end{align}

\subsection{Complementary upper bounds}
\label{app:three_lemmas}

Here we provide the complementary lemmas, and their respective proofs, that complete the proof of Theorem~\ref{thm:upper_WTS_unknown}.

\begin{lemma}
\label{lem:wts_first}
Consider the conditions of Theorem~\ref{thm:upper_WTS_unknown} and the events
\begin{align}
\Asp_t &= \{\norm{\tbmu^\star(t)} \geq \norm{\bmu_1} - \e\},
\notag\\
\Bsp_{k,t} &= \{ \norm{\xx_k(t)-\bmu_k} \leq \e/2 \quad\text{\emph{and}}\quad S_k(t) \leq t(\sigma^2_k + \e) \},
\notag
\end{align}
for every $k=1,\dots,K$ and $t\in\N$, defined for some fixed $\e$ such that $0< \e < \min_{k\neq k^\star} \Delta_k/2$.
Then,
\begin{align}
\sum_{t=3}^T\prob{\tkstar_t = k, \Asp_t,\Bsp_{k,t}}
&\leq
\frac{\sigma_k^2+\e}{(\Delta_k-2\e)^2}\log T + \O(\e^{-2}),
\label{eq:lem1}
\end{align}
for every $k\neq k^\star$, where the $\O(\e^{-2})$ term does not depend on $T$.
\end{lemma}
\begin{proof}
Fix $k$ and define the random variable $z_t \eq \sump$, representing the cumulative power allocated at arm $k$ up to round $t$.
Define also $c(\e)\eq \frac{\sigma_k^2+\e}{(\Delta_k-2\e)^2}>0$, for every $\e>0$ (notice that $c(\e)$ is $\O(1)$ as $\e\to 0$, and independent of $T$).
Then,
\begin{align}
\sum_{t=3}^T \prob{\tkstar_t = k, \Asp_t,\Bsp_{k,t}}
&=
\sum_{t=3}^T \bigg( \prob{\tkstar_t = k, \Asp_t,\Bsp_{k,t},z_{t-1} < c(\e)\log T} 
\notag\\
&\hspace{2.6cm}+ \prob{\tk_t = k, \Asp_t,\Bsp_{k,t}, z_{t-1} \geq c(\e)\log T } \bigg)
\label{eq:twoaddends}
\end{align}
We proceed to upper bound each sum in~\eqref{eq:twoaddends} separately.
Since, for every event $\tilde{E}$, $\prob{\tilde{E}} = \mean{\one{\tilde{E}}} = \mean{\mean{\one{\tilde{E}}\given\Fsp_t}} = \mean{\prob{\tilde{E}\given\Fsp_t}}$,  the first sum satisfies
\begingroup
\allowdisplaybreaks
\begin{align}
\sum_{t=3}^T \prob{\tkstar_t = k, z_{t-1} < c(\e)\log T }
&=
\mean{\sum_{t=3}^T \prob{\tkstar_t = k, z_{t-1} < c(\e)\log T \,\Big|\,\Fsp_t}}
\notag\\
&\overset{(a)}{=}
\mean{\sum_{t=3}^T \one{z_{t-1} < c(\e)\log T} \prob{\tkstar_t = k\given\Fsp_t}  }
\notag\\
&=
\mean{\sum_{t=3}^T p^t_k \one{z_{t-1} < c(\e)\log T }}
\notag\\
&\leq
\mean{\sum_{t=3}^T p^t_k \one{z_{t} < 1 + c(\e)\log T }}
\notag\\
&\leq 1 +  c(\e)\log T
\notag\\
&= \frac{\sigma^2_k+\e}{(\Delta_k-2\e)^2} \log T + \O(1),
\label{eq:clogT}
\end{align}
\endgroup
where $(a)$ follows from $z_{t-1}$ being $\Fsp_t$-measurable and because of Lemma~\ref{lem:divergence}, which guarantees that $\sum_{\ell=1}^t p^\ell_k$ eventually reaches $1 + c(\e)\log T$.
This first bound provides the desired rate of growth in the right-hand side of~\eqref{eq:lem1}, so it remains to prove that the second sum in~\eqref{eq:twoaddends} grows sub-logarithmically, at most, with $T$ large, and polynomially in $\e^{-2}$ as $\e\to 0$.

Deriving an upper bound for the second term in~\eqref{eq:twoaddends} involves bounding the deviation of $\tbmu_k(t)$ around $\xx_k(t-1)$. However, the definition of $\Bsp_{k,t}$ only provides information about $\xx_k(t)$ and $S_k(t)$.
This issue is tackled by formalizing the intuitive property in which $\xx_k(t)$ is \emph{close} to $\xx_k(t-1)$ with high probability.
More precisely, and since the events $\{\tkstar_t = k\}$ and $\Asp_t$ imply
event $\{\norm{\tbmu_k(t)} \geq \norm{\bmu_1}-\e\}$ (which follows from~\eqref{eq:rho_wts}), the second term in~\eqref{eq:twoaddends} satisfies
\begin{align}
&\prob{\tk_t=k,\Asp_t,\Bsp_{k,t}, z_{t-1} \geq c(\e)\log T}
\notag\\
&\quad\leq
\prob{\norm{\tbmu_k(t)}\geq \norm{\bmu_1}-\e,\norm{\xx_k(t)-\bmu_k}\leq \e/2,S_k(t)\leq t(\sigma_k^2 + \e),z_{t-1} \geq c(\e)\log T}
\notag\\
&\quad=
\text{P}\Bigg\{\norm{\tbmu_k(t)}\geq \norm{\bmu_1}-\e,  
\underbrace{\norm{\xx_k(t)-\bmu_k}\leq \e/2, \norm{\xx_k(t)-\xx_k(t-1)}\leq \e/2}_{\implies \norm{\xx_k(t-1)-\bmu_k}\leq \e},
\notag\\
&\quad\hspace{6.8cm}
\underbrace{S_k(t)\leq t(\sigma_k^2 + \e)}_{\implies S_k(t-1)\leq t(\sigma_k^2+\e)},  z_{t-1} \geq c(\e)\log T
\Bigg\}
\notag\\
&\quad\qquad +
\text{P}\Bigg\{\norm{\tbmu_k(t)}\geq \norm{\bmu_1}-\e,  
\underbrace{\norm{\xx_k(t)-\bmu_k}\leq \e/2, \norm{\xx_k(t)-\xx_k(t-1)}\geq \e/2}_{\implies \norm{X^t_k-\bmu_k}\geq z_{t-1}\e/2p^t_k},
\notag\\
&\quad\hspace{7.1cm}
S_k(t)\leq t(\sigma_k^2 + \e),
 z_{t-1} \geq c(\e)\log T\Bigg\}
\notag\\
&\quad\leq
\text{P}\bigg\{\norm{\tbmu_k(t)}\geq \norm{\bmu_1}-\e, \norm{\xx_k(t-1)-\bmu_k}\leq \e,
S_k(t-1)\leq t(\sigma_k^2+\e), z_{t-1} \geq c(\e)\log T 
\bigg\}
\notag\\
&\quad\hspace{2cm}+
\prob{\norm{X^t_k-\bmu_k}\geq z_{t-1}\e/2p^t_k},
\label{eq:some_randome_eq}
\end{align}
which follows from the fact that 
$$\e/2 \leq \norm{\xx_k(t)-\xx_k(t-1)} = \Norm{\xx_k(t) - \frac{z_t\xx_k(t) - p^t_kX^t_k}{z_{t-1}}} =  \frac{p^t_k}{z_{t-1}}\norm{X^t_k-\xx_k(t)}$$
and that $\{S_k(t)\}_t$ is non-decreasing in $t$.
Now, because $\e<\min_{i\neq k^\star} \Delta_i/2 \leq \Delta_k/2$, meaning that $\prob{\norm{\tbmu_k(t)}\geq \norm{\bmu_1}-\e,\norm{\xx_k(t-1)-\bmu_k}\leq \e} \leq \prob{\norm{\tbmu_k(t)-\xx_k(t-1)}\geq \Delta_k-2\e}$, the first term in~\eqref{eq:some_randome_eq} satisfies
\begin{align}
&\text{P}\bigg\{\norm{\tbmu_k(t)}\geq \norm{\bmu_1}-\e, \norm{\xx_k(t-1)-\bmu_k}\leq \e,
S_k(t-1)\leq t(\sigma_k^2+\e), z_{t-1} \geq c(\e)\log T 
\bigg\}
\notag\\
&\hspace{1cm}\leq
\mathbb{E}\Bigg\{\prob{\norm{\tbmu_k(t)-\xx_k(t-1)}\geq \Delta_k - 2\e\given\Fsp_t} 
\notag\\
&\hspace{3.0cm}\times 
\one{\norm{\xx_k(t-1)-\bmu_k}\leq \e,  S_k(t-1)\leq t(\sigma_k^2+\e),  z_{t-1} \geq c(\e)\log T}
\Bigg\}
\notag\\
&\hspace{1cm}=
\mathbb{E}\Bigg\{\left(1+\frac{z_{t-1}(\Delta_k-2\e)^2}{S_k(t-1)}\right)^{-t+3}
\notag\\
&\hspace{3.0cm}\times 
\one{\norm{\xx_k(t-1)-\bmu_k}\leq \e,  S_k(t-1)\leq t(\sigma_k^2+\e),  z_{t-1} \geq c(\e)\log T}
\Bigg\}
\notag\\
&\hspace{1cm}\leq
\mean{\left( 1 + c^{-1}(\e)\frac{z_{t-1}}{t}\right)^{-t+3} \one{z_{t-1} \geq c(\e)\log T}},
\end{align}
while, because $p^t_k$ is $\Fsp_t$-measurable, the second term satisfies
\begin{align}
\sum_{t=3}^T\prob{\norm{X^t_k-\bmu_k}\geq \frac{z_{t-1}\e}{2p^t_k}}
&=
\sum_{t=3}^T\mean{ \prob{\norm{X^t_k-\bmu_k}\geq \frac{z_{t-1}\e}{2p^t_k} \Given\Fsp_t}}
\notag\\
&\overset{(a)}{\leq}
\sum_{t=3}^T\mean{ \exp{-z_{t-1}^2\e^2/4\sigma^2_k p^t_k} }
\notag\\
&\leq
\mean{\sum_{t=3}^T p^t_k \exp{-z_{t-1}^2\e^2/4\sigma^2_k} }
\notag\\
&=
\mean{\sum_{i=0}^T \sum_{t=3}^T  p^t_k \exp{-z_{t-1}^2\e^2/4\sigma^2_k} \one{z_t\in(i,i+1)}}
\notag\\
&\leq 
\sum_{i=1}^T \exp{-(i-1)^2\e^2/4\sigma^2_k} \mean{\sum_{t=3}^T  p^t_k  \one{z_t\in(i,i+1)}  }
\notag\\
&\leq
\exp{\e^2/4\sigma^2_k}\sum_{i=1}^T \exp{-i\e^2/4\sigma^2_k}\,\, 2  
\notag\\
&\leq
2\exp{\frac{\min_{i\neq k^\star}\Delta_i^2}{16\sigma^2_k}}\sum_{i=1}^\infty \exp{-i\e^2/4\sigma^2_k}
\notag\\
&=
2\exp{\frac{\min_{i\neq k^\star}\Delta_i^2}{16\sigma^2_k}} 
\frac{\exp{-\e^2/4\sigma^2_k}}{1-\exp{-\e^2/4\sigma^2_k}}
\notag\\
&= 
\O(\e^{-2}),
\end{align}
where $(a)$ follows from the fact that $1+x \leq \exp{x}$, $x\in\R$, holds, in particular, for $x = -1 + 1/p^t_k$.
In consequence, the second sum in~\eqref{eq:twoaddends} can be upper bounded as
\begingroup
\allowdisplaybreaks
\begin{align}
&\sum_{t=3}^T \prob{\tk_t = k, \Asp_t,\Bsp_{k,t}, z_{t} \geq c(\e)\log T } 
\notag\\
&\hspace{3cm}\leq 
\O(\e^{-2}) + 
\mean{
\sum_{t=3}^T
\left(1 + c^{-1}(\e)\frac{z_{t-1}}{t}\right)^{-t+3} \one{z_t \geq c(\e)\log T}
}
\notag\\
&\hspace{3cm}\leq 
\O(\e^{-2}) + 
(1 + c^{-1}(\e))^3 \,
\mean{\sum_{t=3}^T
\left(1 + \frac{\log T}{t}\right)^{-t} \one{z_t \geq c(\e)\log T}
}
\notag\\
&\hspace{3cm}\leq
\O(\e^{-2}) + 
\left(1+c^{-1}(\e)\right)^3 \,\mean{ \sum_{t\colon z_{t-1}\geq c(\e)\log T}^T \left(1+\frac{\log T}{t-1}\right)^{-t}}
\notag\\
&\hspace{3cm}\leq
\O(\e^{-2}) + 
(1+c^{-1}(\e))^3\sum_{t = \big\lfloor c(\e)\log T\big\rfloor}^T \left(1+\frac{\log T}{t}\right)^{-t}
\notag\\
&\hspace{3cm}\overset{(b)}{\leq}
\O(\e^{-2}) + 
(1+c^{-1}(\e))^3\sum_{t= \big\lfloor c(\e)\log T\big\rfloor}^T \exp{\frac{-t\log T}{t+\log T}}
\notag\\
&\hspace{3cm}\leq
\O(\e^{-2}) + 
(1+c^{-1}(\e))^3 \int_{\big\lfloor c(\e)\log T\big\rfloor-1}^T \exp{\frac{-t\log T}{t+\log T}} \,dt
\notag\\
&\hspace{3cm}\leq
\O(\e^{-2}) + 
(1+c^{-1}(\e))^3 \int_{c(\e)\log(T) -2}^T \exp{\frac{-t\log T}{t+\log T}} \,dt,
\label{eq:1+c}
\end{align}
\endgroup
where $(a)$ follows from the fact that $(1+x/y)^y = \exp{y\log(1+x/y)} \geq \exp{y\frac{x/y}{1+x/y}}$.
Deriving an upper bound for the right-hand side of the equation above involves solving a very delicate integral.
We do this by finding an upper and lower bound of an anti-derivative for $\exp{\frac{-x\log T}{x+\log T}}$.
Such an anti-derivative can be obtained using integration by parts:
\begin{align}
\int \exp{\frac{-x\log T}{x+\log T}} \,dx 
&=
(x+\log T)\exp{\frac{-x\log T}{x+\log T}}  + \int(x+\log T) \exp{\frac{-x\log T}{x+\log T}}\frac{\log^2 T}{(x+\log T)^2}\,dx
\notag\\
&\overset{(a)}{=}
(x+\log T)\exp{\frac{-x\log T}{x+\log T}}  + \int \frac{\log^2 T}{\log(T)-u} \exp{-u}\,du
\notag\\
&\overset{(b)}{=}
(x+\log T)\exp{\frac{-x\log T}{x+\log T}}  - T^{-1} \log^2(T) \int \frac{\exp{v}}{v}\,dv,
\notag
\end{align}
where $(a)$ follows from $u = \log(T)x/(x+\log T)$ and $(b)$ from $v = \log(T)-u$.
The function $f(v) \eq \log v + \sum_{n=1}^\infty \frac{1}{n\cdot n!} v^n$ is an anti-derivative for the function $v\mapsto \exp{v}/v$, since
\begin{align}
\frac{df}{dv}
=
v^{-1} + \sum_{n=1}^\infty \frac{v^{n-1}}{n!}
=
\frac{1}{v}\left(1+\sum_{n=1}^\infty  \frac{v^{n}}{n!} \right)
=
\frac{\exp{v}}{v}.
\end{align}
Additionally, for every $v>0$, it holds that
\begin{align}
\log v \leq f(v) \leq 2\exp{v},
\notag
\end{align}
hence, by reversing the change of variables $v=v(u(x))$, we have that
$$f(v(u(x)) = f\left(\frac{\log^2 T}{x+\log T} \right)$$
is an anti-derivative for $\exp{\frac{-x\log T}{x+\log T}}$ that satisfies
\begin{align}
\log \frac{\log^2 T}{x + \log T} \leq f\left(\frac{\log^2 T}{x + \log T}\right)
\leq
2\exp{ \frac{\log^2 T}{x + \log T}}.
\label{eq:boundEi}
\end{align}
Therefore, the definite integral in the right-hand of~\eqref{eq:1+c} can be upper bounded, by means of~\eqref{eq:boundEi}, as
\begin{align}
&\int_{c(\e)\log T-2}^T \exp{\frac{-t\log T}{t+\log T}} \,dt
\notag\\
&\hspace{1cm}=
(t+\log T)\exp{\frac{-t\log T}{t+\log T}}\Bigg|_{t=c(\e)\log(T)-2}^T  
\!\!\!\!\!\!- T^{-1} \log^2(T)  f\left(\frac{\log^2 T}{t + \log T}\right) \Bigg|_{t=c(\e)\log(T)-2}^T
\notag\\
&\hspace{1cm}=
(T+\log T)\exp{\frac{-T\log T}{T+\log T}} - \left(c(\e)\log(T)-2+\log T\right)\exp{\frac{-\left(c(\e)\log(T)-2\right)\log T}{c(\e)\log(T)-2+\log T}}
\notag\\
&\hspace{1cm}\hspace{2cm}- T^{-1}\log^2(T)  f\left(\frac{\log^2 T}{T + \log T}\right)
+ T^{-1}\log^2 T  f\left(\frac{\log^2 T}{c(\e)\log(T)-2 + \log T}\right)
\notag\\
&\hspace{1cm}\overset{(a)}{\leq}
(T+\log T)\exp{\frac{-T\log T}{T+\log T}} 
- T^{-1}\log^2(T)  f\left(\frac{\log^2 T}{T + \log T}\right)
\notag\\
&\hspace{1cm}\hspace{6.5cm}+ T^{-1}\log^2 T  f\left(\frac{\log^2 T}{c(\e)\log(T)-2 + \log T}\right)
\notag\\
&\hspace{1cm}\leq
2T^{1 - \frac{T}{T+\log T}} 
+ T^{-1}\log^3 (2T)
+ T^{\frac{\log T}{c(\e)\log(T)-2 + \log T}-1} \log^2 T
\notag\\
&\hspace{1cm}\leq 
4 + T^{-1}\log^3(2T) + T^{-1} \log ^2T
\notag\\
&\hspace{1cm}\leq 4 + 2\cdot 3^3/\exp{3} + (2/\exp{})^2
\notag\\
&\hspace{1cm}=
\O(\e^0),
\label{eq:e0}
\end{align}
where $(a)$ follows because the second term in the line above is always negative, since $T$ is assumed to satisfy $c(\e)\log T - 2 > 0$.
The constants hidden in $\O(\e^0)$ are universal (\emph{i.e.}, independent of $T$), therefore, putting together~\eqref{eq:clogT},~\eqref{eq:1+c} and~\eqref{eq:e0} yields
\begin{align}
\sum_{t=3}^T\prob{\tk_t = k,\Asp_t,\Bsp_{k,t}}
&\leq 
\frac{\sigma^2_k+\e}{(\Delta_k-2\e)^2} \log T + \O(1) + \O(\e^{-2}) + \O(1)
\notag\\
&=
\frac{\sigma^2_k+\e}{(\Delta_k-2\e)^2} \log T + \O(\e^{-2}),
\notag
\end{align}
concluding the proof.
\end{proof}


\begin{lemma}
\label{lem:wts_second}
Consider the conditions of Theorem~\ref{thm:upper_WTS_unknown}. Let $\e > 0$, and define the events
\begin{align}
\Bsp_{k,t} &= \{ \norm{\xx_k(t)-\bmu_k} \leq \e/2 \quad\text{\emph{and}}\quad S_k(t) \leq t(\sigma^2_k + \e) \},
\notag
\end{align}
for every $k=1,\dots,K$, and $t\in\N$.
Then,
\begin{align}
\sum_{t=3}^T \prob{\tkstar_t= k,\Bsp_{k,t}^c} = \O(\e^{-2}),
\label{eq:sum_wts_second}
\end{align}
for every $k\neq k^\star$, where the $\O(\e^{-2})$ term is independent of $T$.
\end{lemma}
\begin{proof}
Recall that $h(x) = x - \log(1+x)$.
Fix $k$ and let $z_t\eq \sump$ denote the (random) cumulative allocated power at frequency $k$ up to round $t$.
Define $\Psp_{t+1}$ as the $\sigma$-algebra generated by the power's trajectory up to round $t$, $(p^1,p^2,\dots,p^t)$.
The sum in~\eqref{eq:sum_wts_second} can be decomposed, by means of the union bound, as
\begin{align}
\sum_{t=3}^T \prob{\tkstar_t= k,\Bsp_{k,t}^c} 
&=
\sum_{t=3}^T\prob{\tk_t=k, 
(\norm{\xx_k(t)-\bmu_k} > \e/2 \quad\text{or}\quad  S_k(t) > t(\sigma^2_k + \e))}
\notag\\
&=
\sum_{t=3}^T\prob{\tk_t=k,\norm{\xx_k(t)-\bmu_k} \geq \e/2} 
\notag\\
&\qquad\qquad\qquad
+\sum_{t=3}^T\prob{\tk_t=k, S_k(t) \geq t(\sigma^2_k + \e)},
\label{eq:union_bound_wts_2}
\end{align}
where the second sum in the right-hand side satisfies
\begin{align}
\sum_{t=3}^T\prob{\tk_t=k,S_k(t) \geq t(\sigma^2_k + \e)}
&\leq
\sum_{t=3}^T\prob{S_k(t) \geq t(\sigma^2_k + \e)}
\notag\\
&\leq
\sum_{t=0}^\infty \exp{-th(\e/\sigma_k^2)}
\notag\\
&=
\frac{1}{1-\exp{-h(\e/\sigma_k^2)}}.
\end{align}
On the other hand, because $p^t_k$ is $\Fsp_t$ measurable, meaning that $\sigma(p^1_k,\dots,p^t_k)=\Psp_{t+1}\subset \Fsp_t$, the first sum in the right-hand side of~\eqref{eq:union_bound_wts_2} can be upper bounded as
\begingroup
\allowdisplaybreaks
\begin{align}
&\sum_{t=3}^T \prob{\tk_t=k,\norm{\xx_k(t)-\bmu_k}\geq \e/2}
\notag\\
&\hspace{1cm}=
\mean{\sum_{t=3}^T \prob{\tk_t=k,\norm{\xx_k(t)-\bmu_k}\geq \e/2\Given \Psp_{t+1}}}
\notag\\
&\hspace{1cm}=
\mathbb{E}
\Bigg\{\sum_{i=0}^T\sum_{t=3}^T \one{z_{t}\in(i,i+1)} 
\prob{\norm{\xx_k(t)-\bmu_k}\geq \e/2\ggiven \Psp_{t+1}}
\notag\\
&\hspace{1cm}\hspace{6.0cm}\times
\prob{\tk_t=k\ggiven \Psp_{t+1}, \norm{\xx_k(t)-\bmu_k}\geq \e/2}
\Bigg\}
\notag\\
&\hspace{1cm}=
\mean{\sum_{i=0}^T\sum_{t=3}^T \one{z_{t}\in(i,i+1)} 
\exp{-z_t\e^2/4\sigma_k^2}
\prob{\tk_t=k\ggiven \Psp_{t+1}, \norm{\xx_k(t)-\bmu_k}\geq \e/2}}
\notag\\
&\hspace{1cm}\leq
\sum_{i=0}^T \exp{-i\e^2/4\sigma_k^2}
\sum_{t=3}^T
\mean{
\one{z_{t}\in(i,i+1)} 
\prob{\tk_t=k\ggiven \Psp_{t+1}, \norm{\xx_k(t)-\bmu_k}\geq \e/2}}
\notag\\
&\hspace{1cm}=
\sum_{i=0}^T \exp{-i\e^2/4\sigma_k^2}
\sum_{t=3}^T \mathbb{E}\Bigg\{ \one{z_{t}\in(i,i+1)} 
\notag\\
&\hspace{1cm}\hspace{4.2cm}\times
\mean{
\prob{\tk_t=k\ggiven \Fsp_t, \norm{\xx_k(t)-\bmu_k}\geq \e/2}\ggiven\Psp_{t+1}
}
\Bigg\}
\notag\\
&\hspace{1cm}=
\sum_{i=0}^T \exp{-i\e^2/4\sigma_k^2}
\mean{
\sum_{t=3}^T \one{z_{t}\in(i,i+1)} 
\mean{
p^t_k \ggiven\Psp_{t+1},\norm{\xx_k(t)-\bmu_k}\geq \e/2
}
}
\notag\\
&\hspace{1cm}=
\sum_{i=0}^T \exp{-i\e^2/4\sigma_k^2}
\mean{
\sum_{t=3}^T 
\one{z_{t}\in(i,i+1)}  p^t_k
}
\notag\\
&\hspace{1cm}\leq 
2\sum_{i=0}^\infty \exp{-i\e^2/4\sigma_k^2}
\notag\\
&\hspace{1cm}=
\frac{2}{1-\exp{-e^2/4\sigma_k^2}},
\notag
\end{align}
\endgroup
hence,
\begin{align}
\sum_{t=3}^T \prob{\tkstar_t= k,\Bsp_{k,t}^c} 
&\leq
\frac{2}{1-\exp{-e^2/4\sigma_k^2}} + \frac{1}{1-\exp{-h(\e/\sigma_k^2)}}
\notag\\
&= \O(\e^{-2}) + \O(\e^{-2}),
\end{align}
finishing the proof.
\end{proof}



\begin{lemma}
\label{lem:wts_third}
Consider the conditions of Theorem~\ref{thm:upper_WTS_unknown}. Let  $\e>0$, and define the events
\begin{align}
\Asp_t &= \{\norm{\tbmu^\star(t)} \geq \norm{\bmu_1} - \e\},
\notag
\end{align}
for every  $t\in\N$.
Then,
\begin{align}
\sum_{t=3}^T \prob{\tkstar_t \neq k^\star,\Asp_t^c} = \O(\e^{-4}),
\end{align}
for every arm $k\neq k^\star$, where the $\O(\e^{-4})$ term is independent of $t$.
\end{lemma}
\begin{proof}
Assume, without of generality, that $k^\star = 1$.
Let $z_t\eq \sum_{\ell=1}^t p^t_1$ denote the (random) cumulative power on arm $1$.
Additionally, observe that event $\{\tkstar_t\neq 1,\Asp^c_t\}$ is included in the event $\{\norm{\tbmu_1(t)}\leq \norm{\bmu_1}-\e\}$.
Consider the following chain of inequalities:
\begin{align}
\sum_{t=3}^T \prob{\tkstar_t \neq 1,\Asp_t^c}
&=
\mean{\sum_{t=3}^T \prob{\tk_t\neq 1,\Asp_t^t\given\Fsp_t}}
\notag\\
&\leq
\mean{\sum_{t=3}^T \prob{\norm{\tbmu_1(t)}\leq \norm{\bmu_1}-\e\given\Fsp_t}}
\notag\\
&\leq
\mean{\sum_{t=3}^T \prob{\norm{\tbmu_1(t)-\bmu_1}\geq \e\given\Fsp_t}}
\label{eq:sum_i}
\end{align}
Now, observe that each of the terms in the above equation satisfy
\begin{align}
&\mean{\prob{\norm{\tbmu_1(t)-\bmu_1}\geq \e\given\Fsp_t}}
\notag\\
&\hspace{1cm}=
\mean{\prob{\norm{\tbmu_1(t)-\bmu_1}\geq \e\given\Fsp_t}\one{\norm{\xx_1(t-1)-\bmu_1}> \e/4}}
\notag\\
&\hspace{2.5cm}+
\mathbb{E}\Big\{\prob{\norm{\tbmu_1(t)-\bmu_1}\geq \e\given\Fsp_t}
\notag\\
&\hspace{4cm}\times\one{\norm{\xx_1(t-1)-\bmu_1}\leq \e/4,S_1(t-1)\geq 2(t-1)\sigma_1^2}\Big\}
\notag\\
&\hspace{2.5cm}+
\mathbb{E}\Big\{\prob{\norm{\tbmu_1(t)-\bmu_1}\geq \e\given\Fsp_t}
\notag\\
&\hspace{4cm}\times\one{\norm{\xx_1(t-1)-\bmu_k}\leq \e/4,S_1(t-1) < 2(t-1)\sigma_1^2}\Big\},
\label{eq:three_eqs}
\end{align}
so we now find an upper bound for the above expression by bounding each of its three terms.
The first term satisfies 
\begin{align}
&\mean{\prob{\norm{\tbmu_1(t)-\bmu_1}\geq \e\given\Fsp_t}\one{\norm{\xx_1(t-1)-\bmu_1}> \e/4}}
\notag\\
&\hspace{7cm}\leq
\mean{\mean{\one{\norm{\xx_1(t-1)-\bmu_1}\geq \e/4}\given z_{t-1}}}
\notag\\
&\hspace{7cm}\leq
\mean{\exp{-z_t\e^2/(16\sigma_1^2)}},
\end{align}
which follows from applying Lemma~\ref{lem:conc_ineqs}.
Similarly, the second result in Lemma~\ref{lem:conc_ineqs} implies that 
\begin{align}
&\mathbb{E}\Big\{\prob{\norm{\tbmu_1(t)-\bmu_1}\geq \e\given\Fsp_t}\one{\norm{\xx_1(t-1)-\bmu_1}\geq \e/4,S_1(t-1)\geq 2(t-1)\sigma_1^2}\Big\}
\notag\\
&\hspace{9.5cm}\leq
\prob{S_1(t-1)\geq 2(t-1)\sigma_1^2}
\notag\\
&\hspace{9.5cm}\leq
\exp{-(t-1) h(1)}.
\end{align}
For the third term, we invoke Lemma~\ref{lem:post_WTS}, yielding
\begin{align}
&\prob{\norm{\tbmu_1(t)-\bmu_1}\geq \e\given\Fsp_t}
\notag\\
&\hspace{3cm}=
\int_{\tbmu\colon \norm{\tbmu-\bmu_1}\geq \e}
\frac{z_{t-1}(t-3)}{\pi S_1(t-1)}
\left(1 + \frac{z_{t-1}\norm{\tbmu-\xx_1(t-1)}^2}{S_1(t-1)}\right)^{-t+2} \!\!\!\!\!\!d\tbmu
\notag\\
&\hspace{3cm}\overset{(a)}{\leq}
\int_{\tbmu\colon \norm{\tbmu-\xx_1(t-1)}\geq \e/2}
\frac{z_{t-1}(t-3)}{\pi S_1(t-1)}
\left(1 + \frac{z_{t-1}\norm{\tbmu-\xx_1(t-1)}^2}{S_1(t-1)}\right)^{-t+2} \!\!\!\!\!\!d\tbmu
\notag\\
&\hspace{3cm}=
\int_{0}^{2\pi}\int_{\e/2}^\infty \frac{z_{t-1}(t-3)}{\pi S_1(t-1)}
\left(1 + \frac{z_{t-1}r^2}{S_1(t-1)}\right)^{-t+2} r\,dr\,d\theta
\notag\\
&\hspace{3cm}=
2\left(1+\frac{z_{t-1}\e^2}{4S_1(t-1)}\right)^{-t+3}
\notag\\
&\hspace{3cm}\overset{(b)}{\leq}
2\left(1+\frac{z_{t-1}\e^2}{8(t-1)\sigma^2}\right)^{-t+3},
\end{align}
where $(a)$ follows from the fact that $\xx_1(t-1)$ is $\e/4$ close to $\bmu_1$, meaning that one can always construct a circle of radius $\e/2$ around $\xx_1(t-1)$ (that encircles $\bmu_1$) that is completely contained inside $\{\tbmu\colon\norm{\tbmu-\bmu_1}\geq \e\}$.
The inequality in $(b)$ holds because the sample variance is restricted to $S_1(t-1)\leq 2(t-1)\sigma_k^2$.
Putting these three bounds together leads to
\begin{align}
&\mean{\prob{\norm{\tbmu_1(t)-\bmu_1}\geq\e\given\Fsp_t}}
\notag\\
&\hspace{4.0cm}\leq
\mean{
\exp{-z_{t-1}\e^2/(16\sigma_1^2)} + \exp{-(t-1)h(1)} + 2\left(1+\frac{z_{t-1}\e^2}{8(t-1)\sigma_1^2}\right)^{-t+3}},
\end{align}
where the expectation is respect to the sequence $z_{t-1}$.
Then, the expected sum in~\eqref{eq:sum_i} satisfies
\begin{align}
&\sum_{t=3}^T \prob{\tkstar_t \neq k^\star,\Asp_t^c}
\notag\\
&\hspace{1cm}\leq
\mean{\sum_{t=3}^{\infty} \exp{-z_{t-1}\e^2/(16\sigma_1^2)} + \exp{-(t-1)h(1)} + 2\left(1+\frac{z_{t-1}\e^2}{8(t-1)\sigma_1^2}\right)^{-t+3}}
\notag\\
&\hspace{1cm}=
\mean{\sum_{t=2}^{\infty} \exp{-z_{t}\e^2/(16\sigma_1^2)} + \exp{-th(1)} + 2\left(1+\frac{z_{t}\e^2}{8t\sigma_1^2}\right)^{-t+2}}
\notag\\
&\hspace{1cm}\overset{(a)}{\leq}
\mean{\sum_{i=0}^\infty \sum_{z_t\in(i,i+1)} 
\left( \exp{-z_{t}\e^2/(16\sigma_1^2)} + 2\left(1+\frac{\e^2}{8\sigma_1^2}\right) \exp{-z_t\e^2/( \e^2+8\sigma_1^2)}\right)}
+ \frac{1}{1-\exp{-h(1)}}
\notag\\
&\hspace{1cm}\leq
\mean{\sum_{i=0}^\infty \left( \exp{-i\e^2/(16\sigma_1^2)} + 2\left(1+\frac{\e^2}{8\sigma_1^2}\right) \exp{-i\e^2/(\e^2+8\sigma_1^2)}\right) \sum_{z_t \in(i,i+1)} 1}
+ \frac{1}{1-\exp{-h(1)}}
\notag\\
&\hspace{1cm}=
\sum_{i=0}^\infty \left( \exp{-i\e^2/(16\sigma_1^2)} + 2\left(1+\frac{\e^2}{8\sigma_1^2}\right) \exp{-i\e^2/(\e^2+8\sigma_1^2)} \right) \mean{\sum_{z_t \in(i,i+1)} 1}
+ \frac{1}{1-\exp{-h(1)}}
\notag\\
&\hspace{1cm}\overset{(b)}{\leq}
\sum_{i=0}^\infty \left( \exp{-i\e^2/(16\sigma_1^2)} + 2\left(1+\frac{\e^2}{8\sigma_1^2}\right) \exp{-i\e^2/(\e^2+8\sigma_1^2)} \right) C_{\bmu,\bsigma} 
+ \frac{1}{1-\exp{-h(1)}}
\notag\\
&\hspace{1cm}=
C_{\bmu,\bsigma} \left(
\frac{1}{1-\exp{-\e^2/(16\sigma^2_1)}} + 2\left(1+\frac{\e^2}{8\sigma_1^2}\right) \frac{1}{1-\exp{-\e^2/(\e^2+8\sigma_1^2)}} \right) 
+ \frac{1}{1-\exp{-h(1)}}
\notag\\
&\hspace{1cm}=
\O(\e^{-2}) + \O(\e^{-2}) + \O(\e^{-2})
\notag\\
&\hspace{1cm}=
\O(\e^{-2}),
\end{align}
where $h(1) = 1-\log 2>0$.
Step $(a)$ is obtained from the inequality $(1+x/y)^{-y} = \exp{-y\log(1+x/y)} \leq \exp{-y \frac{x/y}{1+x/y}}$, with $x=z_t\e^2/(8\sigma_1^2)$ and $y=t$, so
$$\expo{\frac{-t\frac{z_t\e^2}{8\sigma_1^2}}{t + \frac{z_t\e^2}{8\sigma_1^2}}} 
\leq 
\expo{\frac{-t\frac{z_t\e^2}{8\sigma_1^2}}{t\left(1 + \frac{1\e^2}{8\sigma_1^2}\right)}},
$$
because $z_t\leq t$.
Inequality $(b)$ is a very crucial part of this result that deserves its own statement and proof, which can be found in Lemma~\ref{lem:bound_sum_z}.
The importance of that result is that is allows us to upper bound the expected number of rounds it takes to apply a cumulative power exceeding 1 at arm $k^\star$.
Such a bound is problem-dependent constant $C_{\bmu,\bsigma}$ but it does not depend on $t$ nor $\e$.
\end{proof}

\subsection{Technical Lemmas}
\label{app:technical_lemmas}

In this section we provide three technical lemmas that support the ones stated in Section~\ref{app:three_lemmas}:
\begin{itemize}
\item Lemma~\ref{lem:conc_ineqs} provides concentration inequalities (and equalities) for the sufficient statistics $\xx_k(t)$ and $S_k(t)$, the posterior means $\tbmu_k(t)$.
\item Lemma~\ref{lem:divergence} proves a key property of \WTS, in which the cumulative power allocated at every arm diverges to $\infty$ with probability 1.
This is, in fact, a very desirable property since learning the optimal arm involves knowing $\bmu$ completely which, by the law of large numbers~\cite{papoul91}, is only achievable by sampling all arms an infinite number of times or, in our case, with an infinite cumulative power.
\item Lemma~\ref{lem:bound_sum_z} bounds the minimum number of rounds required to bring the cumulative power at arm 1 from $i\in\N$ to $i+1$ when the power profiles are determined by WTS (see Algorithm~\ref{alg:wts}).
Such a bound is provided in the form of a problem-dependent constant, independent of $T$ and $\e$.
This lemma is invoked only in the proof of Lemma~\ref{lem:wts_third}.
\end{itemize}

\begin{lemma}
\label{lem:conc_ineqs}
For every $k\in\setK$, $t\in\setT$, and $\e>0$, it holds that
\begin{align}
\prob{\norm{\xx_k(t) - \bmu_k}+\e\given p^1_k,p^2_k,\dots,p^t_k} &= \exp{\frac{-\sump}{\sigma_k^2}\e^2},
\end{align}
and
\begin{align}
\prob{S_k(t)\geq t(\sigma_k^2+\e) } &\leq 
\exp{-th(\e/\sigma_k^2)},
\label{eq:boundS}
\end{align}
where $h(x)\eq x-\log(1+x)$, $\forall x>0$.
Furthermore,
\begin{align}
\prob{\norm{\tbmu_k(t)-\xx_k(t-1)}\geq\delta\given \Fsp_t}
&=
\left( 1 + \frac{\sum_{\ell=1}^{t-1}p^\ell_k \delta^2}{S_k(t-1)}\right)^{-t+3}.
\end{align}
\end{lemma}

\begin{proof}
Let $\Psp_{t+1}$ denote the sigma algebra generated by $p^1_k,p^2_k,\dots,p^t_k$.
By Lemma~\ref{lem:WTS_suff}, we have that
\begin{align}
\prob{\norm{\xx_k(t)-\bmu_k}\geq \e \given\Psp_{t+1} }
&=
\int_{\bx:\norm{\bx-\bmu_k}\geq \e} \frac{\sump}{\pi\sigma_k^2}\exp{\frac{-\sump}{\sigma_k^2}\norm{\bx-\bmu_k}^2}\,d\bx
\notag\\
&=
\int_{\bz:\norm{\bz}\geq \e} \frac{\sump}{\pi\sigma_k^2}\exp{\frac{-\sump}{\sigma_k^2}\norm{\bz}^2}\,d\bz
\notag\\
&=
\int_0^{2\pi}\int_\e^\infty \frac{\sump}{\pi\sigma_k^2}\exp{\frac{-\sump}{\sigma_k^2}r^2}r\,dr\,d\theta
\notag\\
&=
\exp{\frac{-\sump}{\sigma_k^2}\e^2},
\end{align}
which follows from the change of variables $\bz \eq \bx-\bmu_k$, followed by a polar change of variable.

To derive~\eqref{eq:boundS}, we relax Chernoff's bound on $S_k(t)/(\sigma_k^2/2)\sim\chi^2_{2(t-1)}$:
\begin{align}
\prob{S_k(t)\geq t(\sigma_k^2+\e)}
&\leq
\exp{\inf_{\lambda<1/\sigma_k^2}\log\mean{\exp{\lambda S_k(t)}}-\lambda t(\sigma_k^2+\e)}
\notag\\
&=
\left(\frac{t}{t-1}\right)^{t-1}\left(1+\frac{\e}{\sigma^2_k}\right)^{t-1}\exp{-(t\e/\sigma_k^2+1)}
\notag\\
&=
\left(\frac{t}{t-1}\right)^{t-1}\left(1+\frac{\e}{\sigma^2_k}\right)^{-1}\exp{t\log\left(1+\e^2/\sigma_k^2\right)}\exp{-(t\e/\sigma_k^2+1)}
\notag\\
&=
\left(\frac{t}{t-1}\right)^{t-1}\exp{-1}\underbrace{\left(\frac{\sigma_k^2}{\sigma_k^2+\e^2}\right)}_{\leq 1} \exp{-t\left(\e^2/\sigma_k^2-\log\left(1+\e^2/\sigma_k^2\right)\right)}
\notag\\
&\leq
\exp{-t\,h(\e/\sigma_k^2)},
\end{align}
since $\left(\frac{t}{t-1}\right)^{t-1}\leq \exp{1}$ for every $t>1$.

For the last part of the proof, observe that, by Lemma~\ref{lem:post_WTS}, and by changing to polar coordinates, we have that
\begin{align}
&\prob{\norm{\tbmu_k(t)-\xx_k(t-1)}\geq\delta\given\Fsp_t}
\notag\\
&\hspace{2.5cm}=
\int_{\tbmu:\norm{\tbmu-\xx_k(t\text{-}1)}>\delta} \!\!\!\!
\frac{\sum_{\ell=1}^{t-1}p^\ell_k(t-3)}{\pi S_k(t-1)}\left( 1 + \frac{\sum_{\ell=1}^{t-1}p^\ell_k\norm{\tbmu-\xx_k(t-1)}^2}{S_k(t-1)}\right)^{-t+2} 
\!\!\!\!\!\!\!d\tbmu
\notag\\
&\hspace{2.5cm}=
\int_{\bz:\norm{\bz}>\delta} \frac{\sum_{\ell=1}^{t-1}p^\ell_k(t-3)}{\pi S_k(t-1)}\left( 1 + \frac{\sum_{\ell=1}^{t-1}p^\ell_k \norm{\bz}^2}{S_k(t-1)}\right)^{-t+2} d\bz
\notag\\
&\hspace{2.5cm}=
\int_0^{2\pi}\int_\delta^\infty \frac{\sum_{\ell=1}^{t-1}p^\ell_k(t-3)}{\pi S_k(t-1)}\left( 1 + \frac{\sum_{\ell=1}^{t-1}p^\ell_k r^2}{S_k(t-1)}\right)^{-t+2}r\,dr\,d\theta
\notag\\
&\hspace{2.5cm}=
\left( 1 + \frac{\sum_{\ell=1}^{t-1}p^\ell_k\delta^2}{S_k(t-1)}\right)^{-t+3}.
\end{align}
finishing the proof.
\end{proof}

Lemma~\ref{lem:conc_ineqs} shows that, as in  \TS~\cite{mulroj19}, the posterior means are symmetrically distributed around the sample (empirical) mean.
It also illustrates that the sample means ($\xx_k(t)$) concentrate around their respective means ($\bmu_k$) at different speeds, depending on which algorithm is being used: in \TS the sample mean of arm $k$ concentrates at a rate $\propto \exp{-N_k(t)\e^2}$, with $N_k(t)$ being the number of times arm $k$ has been sampled up to round $t$, while in \WTS, it concentrates at a rate $\propto\exp{-\sum_{\ell=1}^t p^{\ell}_k\e^2}$ (compare this Lemma~\ref{lem:conc_ineqs} to Lemmas 3 and 4 in~\cite{mulroj19}).
On the other hand, the sample variances $S_k(t)$ concentrate exponentially in $t$ around $t\sigma_k^2$, regardless of the algorithm (\TS or \WTS).
The concentration speed of $S_k(t)$ is consequence of two facts: (i) the outcome from arm $k$ provides enough information about $\sigma_k^2$ no matter how small $p^t_k$ is, and (ii) albeit small, \WTS constantly assigns resources to all arms, as the following lemma shows.
It is the imbalance in the concentration rates of the sample mean and variance under \WTS what introduces difficulties when trying to prove that \WTS is an optimal algorithm.

\medskip
\begin{lemma}
\label{lem:divergence}
Consider problem~\emph{(RM)} under unknown variance $\bsigma$.
Then, the cumulative power at arm $k$ satisfies
\begin{align}
\lim_{T\to\infty} \sum_{t=1}^T p^t_k = \infty,\quad\text{a.s.}
\end{align}
for every $k=1,\dots,K$.
\end{lemma}
\begin{proof}
The sequence $(\sump)_t$ is non-decreasing, since $p^t_k\geq 0$ for every $t=1,\dots,T$.
Furthermore, $p^t_k$ is never equal to zero since the posterior means have infinite support.
The proof is by contradiction.
Fix $k\in\setK$ and assume $\lim_{T\to\infty}\sum_{t=1}^T p^t_k = c <\infty$.
Then, by Lemmas~\ref{lem:WTS_suff} and~\ref{lem:conc_ineqs}, $\xx_k(t)$ and $S_k(t)/t$ converge to some (random) $\xx_k\eq \lim_{t\to\infty}\xx_k(t)\sim\Nsp(\bmu_k,\sigma_k^2/(2c)\bI_2)$ and $\lim_{t\to\infty}S_k(t)/t = \sigma_k^2$ (almost surely).
From Lemma~\ref{lem:post_WTS}, we have that the limit posterior mean distribution is
\begin{align}
\lim_{t\to\infty} f_{\bmu\given\Fsp_t}(\tbmu)
&=
\lim_{t\to\infty} \frac{\sum_{\ell=1}^{t-1}(t-3)}{\pi S_k(t-1)}\left( 1+ \frac{\sum_{\ell=1}^{t-1}\norm{\tbmu-\xx_k(t-1)}^2}{S_k(t-1)}\right)^{-t+2}
\notag\\
&=
\frac{c}{\sigma^2_k \pi}\exp{-c\norm{\tbmu-\xx_k} ^2/\sigma_k^2},
\quad\text{a.s.}
\end{align}
thus, given that $\xx_k\neq\tbmu$ (which has measure zero), we have that $\lim_{t\to\infty} f_{\bmu_k\given\Fsp_t}$ defines a non-degenerate distribution of infinite support and, therefore, $\lim_{t\to\infty}\prob{\cap_{i\neq k}\{\norm{\tbmu_k(t)}\geq \norm{\tbmu_i(t)}\}\given\Fsp_t} = c' \neq 0$, for some $c' = c'(\bmu,\bsigma)>0$ (provided such limits exist). 
We then conclude that $\lim_{t\to\infty}p^t_k = c' > 0$, implying that the sequence $(p^\ell_k)_\ell$ does not converge, so neither does $\sum_{\ell=1}^t p^\ell_k$, leading to a contradiction. This concludes the proof.
\end{proof}

\begin{lemma}
\label{lem:bound_sum_z}
Consider $t\geq 3$.
The expected number of rounds in which the cumulative power at arm $k^\star=1$, $z_t \eq  \sum_{\ell=1}^t p^t_1$, belongs to the set $(i,i+1)$ satisfies
\begin{align}
\mean{\sum_{t\colon z_t\in(i,i+1)} 1} 
\leq 
C_{\bmu,\bsigma}
\eq
2\left(1-\exp{\frac{3}{2K}\left(\frac{\Delta_{\min}}{\sigma_{\max}}\right)^2}\right)^{-2K} \left(1-\exp{-2h(1)}\right)^{-K},
\notag
\end{align}
where $\Delta_{\min}\eq \min_k\Delta_k$ and $\sigma_{\max} \eq \max_{k}\sigma_k$
that is, $C_{\bmu,\bsigma}$ is a problem-dependent constant, independent of $t$.
\end{lemma}

\begin{proof}
Define $z^t_k \eq \sum_{\ell=1}^t p^t_k$, for every $k=1,\dots, K$, and observe that $z_t = z^t_1$.
Proving this result involves firstly lower bounding the expected value of $p^{t+1}$ given $z_{t}$.
Recall that $p^{t+1}_1 = \prob{\tk_{t+1}=1}$.
Now, observe that the event $\cap_{k=1}^K \{ \norm{\xx_k(t)-\bmu_k} \leq \Delta_{\min}, \norm{\tbmu_k(t+1)-\xx_k(t)} \leq \Delta_{\min}\}$ implies that $\tk_{t+1}=1$.
In words, the event in which the sample means are close to their respective true means, together with the event in which the posteriors are close to their respective means (the sample means), all of them very likely to happen, imply that $\tk_{t+1}=1$.
Observe that $\tk_{t+1}=1$ holds regardless of the values of $\{S_k(t)\}_{k,t}$ and, in particular, of the event $\cap_{k=1}^K \{S_k(t)\leq 2t\sigma_k^2\}$ (which is also likely), so it holds that
\begin{align}
&\mean{p^{t+1}_1\given z_t}\hspace{1cm}
\notag\\
&\hspace{1cm}=
\mean{\mean{\one{\tk_{t+1} = 1}\given z_t}\given \Fsp_t}
\notag\\
&\hspace{1cm}=
\prob{\tk_{t+1}=1\given z_t}
\notag\\
&\hspace{1cm}\geq
\prob{\bigcap_{k=1}^K\{\norm{\xx_k(t)-\bmu_k}\leq\Delta_{\min}, S_k(t)\leq 2t\sigma_k^2,\norm{\tbmu_k(t+1)-\xx_k(t)}\leq \Delta_{\min}\}\ggiven z_t}
\notag\\
&\hspace{1cm}\geq
\mean{\!\oneX{\bigcap_{k=1}^K\{\norm{\xx_k(t)-\bmu_k}\!\leq\!\Delta_{\min}, S_k(t)\!\leq\! 2t\sigma_k^2,\norm{\tbmu_k(t+1)-\xx_k(t)}\!\leq\!\Delta_{\min}\}\!}\!\!\ggiven z_t\!}
\notag\\
&\hspace{1cm}\overset{(a)}{=}
\mathbb{E}\Bigg\{\!\oneX{\bigcap_{k=1}^K\{\norm{\xx_k(t)-\bmu_k}\!\leq\!\Delta_{\min}, S_k(t)\!\leq \! 2t\sigma_k^2\}\!} 
\!\prod_{k=1}^K\! \left(\!1\!-\!\left(\!1\!+\!\frac{z^t_k\Delta_{\min}^2}{S_k(t)} \right)^{\!-t+3}\right)\!\!
\ggiven z_t\!\Bigg\}
\notag\\
&\hspace{1cm}\overset{(b)}{\geq}
\left(1- \exp{-\frac{3}{2K}\left(\frac{\Delta_{\min}}{\sigma_{\max}}\right)^2}\right)^{K}
\mathbb{E}\Bigg\{\one{\bigcap_{k=1}^K\{\norm{\xx_k(t)-\bmu_k}\leq\Delta_{\min}, S_k(t)\leq 2t\sigma_k^2\}} 
\ggiven z_t\Bigg\}
\notag\\
&\hspace{1cm}\overset{(c)}{=}
\left(1- \exp{-\frac{3}{2K}\left(\frac{\Delta_{\min}}{\sigma_{\max}}\right)^2}\right)^{K}
\prod_{k=1}^K \left(1-\exp{-z_t\Delta_{\min}^2/\sigma_k^2}\right)
\left( 1-\exp{-th(1)} \right)
\notag\\
&\hspace{1cm}\geq
\left(1- \exp{-\frac{3}{2K}\left(\frac{\Delta_{\min}}{\sigma_{\max}}\right)^2}\right)^{2K}
\left( 1-\exp{-2h(1)} \right)^K
\notag\\
&\hspace{1cm}\qe 
c_{\bmu,\bsigma},
\label{eq:c_small}
\end{align}
where the equalities in $(a)$ and $(c)$ follow from the tower property of expectations with $\sigma(z_t)\subset \Fsp_t$ and  $\sigma(z_t)\subset \sigma(z_2^t,z_3^t,\dots,z_K^t)$, respectively.
The inequalities are a consequence of invoking Lemma~\ref{lem:conc_ineqs}, and the inequality $(b)$ follows because \WTS enforces that the first two power profiles $p^1,p^2$ satisfy $p^t_k = K^{-1}$, for every arm $k=1,\dots,K$ (see Algorithm~\ref{alg:wts}).

Fix $i\in\N$ and let us introduce $\ubar{t}\eq\min\{t\in\N\colon z_t \geq i\}$ and $\bar{t}\eq\min\{t>\ubar{t}\colon z_t \geq i+1\}$.
Observe that $\bar{t}$ denotes a stopping time in $\sigma(z_1,z_2,\dots,z_t)$.
Define also the partial sums $\xi_t \eq \sum_{n=\ubar{t}}^{\bar{t}} p^n_1.$
Then,
\begingroup
\allowdisplaybreaks
\begin{align}
\mean{\xi_{\bar{t}}}
&=
\sum_{t=\ubar{t}}^\infty \mean{\xi_t\one{\bar{t}=t}}
\notag\\
&=\sum_{t=\ubar{t}}^\infty \sum_{n=\ubar{t}}^t \mean{p_1^n \one{\bar{t}=t}}
\notag\\
&=
\sum_{n=\ubar{t}}^\infty \mean{p_1^n \one{\bar{t}\geq n }}
\notag\\
&\overset{(a)}{=}
\sum_{n=\ubar{t}}^\infty \mean{\prob{\tk_n = 1\given z_{n-1}}\one{\bar{t}\geq n}}
\notag\\
&\overset{(b)}{\geq}
c_{\bmu,\bsigma}\sum_{n=\ubar{t}}^\infty \prob{\bar{t}\geq n}
\notag\\
&=
c_{\bmu,\bsigma}\sum_{n=0}^\infty\prob{\bar{t}-\ubar{t}\geq n}
\notag\\
&=
c_{\bmu,\bsigma}\mean{\bar{t}-\ubar{t}},
\end{align}
\endgroup
where in step $(a)$ we used the fact that $\{\bar{t}<n\}^c = \{\bar{t}\geq n\}$ is $\sigma(z_1,z_2,\dots,z_{n-1})$-measurable, and $(b)$ is a consequence of~\eqref{eq:c_small}.
Now notice that the partial sums satisfy $1\leq\xi_{\bar{t}}\leq 2$, which allows us to conclude that 
$
\mean{\sum_{t\colon z_t\in(i,i+1)} 1} 
=
\mean{\bar{t}-\ubar{t}}
\leq
c_{\bmu,\bsigma}^{-1}\mean{\xi_{\bar{t}}}
\leq
2c_{\bmu,\bsigma}^{-1}
\qe
C_{\bmu,\bsigma},
$
concluding the proof.
\end{proof}

%% file: contents/Appendix/rho.tex
\label{app:rho}
The goal of this section is to provide with a method that allows us to update $\rho^t$ in Algorithm~\ref{alg:wts} in a tractable manner.
Obtaining $\rho^t =$ Prob$\{\tk_t=k\given\Fsp_t\}$ can be carried out indirectly by first finding the distributions $(f_{\bmu_k\given\Fsp_t})_{k=1}^K$, as explained in~\eqref{eq:rho_wts}.
However, finding a closed form for the mapping from $(f_{\bmu_k\given\Fsp_t})_{k=1}^K$ to $\rho^t$ involves the calculation of $K$-dimensional very complicated integrals:
\begin{align}
\rho_k^t = \int_{(\tbmu_1,\dots,\tbmu_K)\in\R^{2K}} 
\hspace{-0.5cm}\one{\hspace{-2pt}\bigcap_{i\neq k} \{\norm{\tbmu_i}\leq\norm{\tbmu_k}\}\hspace{-3pt}} f_{\bmu_1\given\Fsp_t}(\tbmu_1)\dots f_{\bmu_K\given\Fsp_t}(\tbmu_K)d(\tbmu_1,\dots,\tbmu_K),
\notag
\end{align}
for which no closed-form solution exists.
We overcome this practical issue by estimating $\rho^t$ through the Monte Carlo method~\cite{fishma96} defined in Algorithm~\ref{alg:mc}.

\begin{algorithm}
\caption{Estimation of $\rho^t$}
\label{alg:mc}
\begin{algorithmic}[1]
      	\State Input: $(f_{\bmu_1\given\Fsp_t}, f_{\bmu_2\given\Fsp_t},\dots, f_{\bmu_K\given\Fsp_t})$ (posterior distributions at round $t$), $M$ (number of samples per arm)
      	\State Draw $M$ samples $\tbmu^{(1)}_k, \tbmu^{(2)}_k,\dots,\tbmu^{(M)}_k \sim f_{\bmu_k\given\Fsp_t}$, for every $k=1,\dots,K$
      	\For{$k=1$ to $K$}
        	\State Set $\rho^t_k \approx \frac{1}{M} \sum_{m=1}^M \one{        			\arg\max_i\{\norm{\tbmu^{(m)}_i}\} = k}$
        \EndFor
\end{algorithmic}
\end{algorithm}
Observe that, in Algorithm~\ref{alg:mc}, $\frac{1}{M} \sum_{m=1}^M \onesmall{\arg\max_i\{\norm{\tbmu^{(m)}_i}\} = k} \overset{\text{a.s.}}{\to} \rho^t_k$ as $M\to\infty$, so picking $M$ large ensures a good approximation of $\rho^t$.
Observe also that \WTS in Algorithm~\ref{alg:wts} coincides with \TS when $\rho^t$ is updated using Algorithm~\ref{alg:mc} with $M=1$.

For completeness, we explain here how to sample from $\tbmu_k(t)\sim f_{\bmu_k\given\Fsp_t}$, for every $k=1,\dots,K$, and $t=4,5,\dots$.
Fix $k$ and $t$, and recall that, from Lemma~\ref{lem:post_WTS}, the posterior means $\tbmu_k(t)$ satisfy
\begin{align}
f_{\bmu_k\given\Fsp_t}(\tbmu)
&=
\frac{\sum_{\ell=1}^{t-1}p^\ell_k(t-3)}{\pi S_k(t-1)}
\left( 1 + \frac{\sum_{\ell=1}^{t-1}p^\ell_k\norm{\tbmu-\xx_k(t-1)}^2}{S_k(t-1)}\right)^{-t+2}.
\notag
\end{align}
Now, consider the following change of variables into polar coordinates $\tbmu_k(t) = \xx_k(t-1) + r[\cos\theta\quad\sin\theta]^\top$.
Then, the distribution of the new variables $(r,\theta)$ is~\cite{papoul91}
\begin{align}
f_{r,\theta}(r,\theta)
&=
r\,f_{\bmu_k\given\Fsp_t}\left(\xx_k(t-1) + r\begin{bmatrix}
\cos\theta\\ \sin\theta
\end{bmatrix}\right)
\notag\\
&=
\underbrace{\frac{1}{2\pi}}_{\qe f_\theta(\theta)}
\underbrace{
\frac{2 r \sum_{\ell=1}^{t-1}p^\ell_k(t-3)}{\pi S_k(t-1)}
\left( 1 + \frac{\sum_{\ell=1}^{t-1}p^\ell_k r^2}{S_k(t-1)}\right)^{-t+2}
}_{\qe f_r(r)},
\notag
\end{align}
meaning that $\theta$ and $r$ are statistically independent random variables that can be sampled separately.
Observe that $\theta$ is uniformly distributed over $[0,2\pi]$.
On the other hand, sampling $r$ can be achieved by means of the inverse sampling theorem.
To this end, let $F(r)$ denote the cumulative density function of $r$:
\begin{align}
F(r) = \int_0^r f_r(u) du
=
1 - \left( 1 + \frac{\sum_{\ell=1}^{t-1}p^\ell_k r^2}{S_k(t-1)}\right)^{-t+3},
\notag
\end{align}
meaning that the inverse $F^{-1}$ is defined by
\begin{align}
r = F^{-1}(\xi) = \sqrt{\frac{S_{t-1}}{\sum_{\ell=1}^{t-1}p^\ell_k} \left( (1-\xi)^{\frac{1}{-t+3}}-1\right)},
\notag
\end{align}
for every $\xi\in[0,1]$ because the image of $F$ is $[0,1]$.
Therefore, a sample from $r\sim f_r$ can be obtained by first sampling from $\xi\sim\Usp([0,1])$ (where $\Usp$ stands for uniform distribution) and then obtaining $r = F^{-1}(\xi)$~\cite{papoul91}.
In consequence, a sample from $\tbmu_k(t)\sim f_{\bmu_k\given\Fsp_t}$ can be obtained as $\tbmu_k(t) = \xx_k(t-1) + r[\cos\theta\quad\sin\theta]^\top$.

%% file: contents/Appendix/simulation.tex
\label{app:simulation}

\begin{figure}[t]
\centering 
\def\svgwidth{1\textwidth}
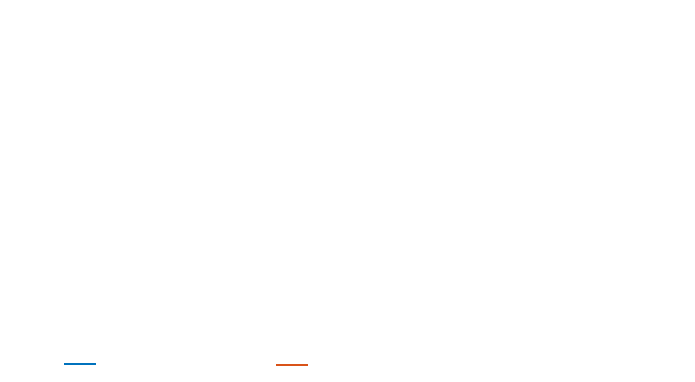 
\caption{The values of $\bmu$ and $\bsigma$ used in the simulation study in Section~\ref{subsec:WTS_sim} based on the frequency responses of $G$ and $H$.
The two-dimensional means and the variances satisfy $\norm{\bmu_k} = |G(\ejwk)|$ while $\sigma^2_k = |H(\ejw)|^2$, respectively.
}
\label{fig:WTS_sim}
\end{figure}

In this section we provide more details on the values of $(K,\bmu,\bsigma)$ in the simulation study in Section~\ref{subsec:WTS_sim}.
Since the problem formulated in this work is inspired by the gain estimation problem of an LTI system, we use LTI filters $G$ and $H$ to define $\bmu$ and $\bsigma$ from their frequency responses in Fig.~\ref{fig:WTS_sim}.
We choose $K=10$ in order to have noticeably different lower bounds (see Fig.~\ref{fig:all_algs}) and where $\bmu_k = |G(\ejwk)|$ for $\wk = 2\pi/(2K+1)$, $k=1,\dots,K$.
On the other hand, the variance of the experiments is set to $\sigma_k^2 = |H(\ejwk)|^2$, for every $k=1,\dots,K$.
The value of $\rho^t$ in \WTS (see Algorithm~\ref{alg:wts}) is updated via Algorithm~2 in Appendix~\ref{app:rho} with $M=500$ Monte Carlo simulations.

\begin{algorithm}
\caption{\quad \WTS under known $\bsigma$~\cite{muller17}}
\label{alg:wts_known}
\begin{algorithmic}[1]
      	\State Input: $T$, $\rho^1 = (1/K, 1/K,\dots,1/K)$ (prior distribution for $\tk_1$), $\lambda^2$, $M$
		\For{$t = 1$ to $T$}
            \State Apply the power profile $p^{t,\WTS} = \rho^1$ and collect the outcome $X^t$
            \State Obtain $f_{\bmu_k\given\Fsp_{t+1}} = \Nsp(\bm^{t+1}_k,v^{t+1}_k\bI_2)$ for every $k=1,\dots,K$
            \begin{align}
            \bm^{t+1}_k &= \frac{\lambda^2 \sum_{\ell=1}^t p^{\ell}_k X^\ell_k}
            				{\sigma^2_k + \lambda^2\sum_{\ell=1}^t p^{\ell}_k}
            \notag\\
            v^t_k &= \frac{\sigma_k^2\lambda^2}{1+\lambda^2\sum_{\ell=1}^t p^{\ell}_k}
            \notag
            \end{align}
            \State Update $\rho^{t+1}$ via Algorithm~\ref{alg:mc}
        \EndFor
\end{algorithmic}
\end{algorithm}

We implemented four algorithms:
\begin{itemize}
\item \TS under known $\bsigma$ based on the algorithm introduced in~\cite{muller17},
\item \WTS under known $\bsigma$ based on the algorithm introduced in~\cite{muller17},
\item \TS under unknown $\bsigma$ based on the algorithm described in~\cite{mulroj19}, and
\item \WTS as proposed in Algorithm~\ref{alg:wts} in this work.
\end{itemize}

It is important to mention that \TS and \WTS under known variance are initialized with Gaussian prior distributions for every mean $\bmu_k$ of the form $f_{\bmu_k\given\Fsp_1} = \Nsp(\bzero, \lambda^2 \bI)$, where $\lambda=1$ is arbitrarily chosen.
Different values of $\lambda$ may improve (or worsen) the finite-time regrets (or the regret in a particular episode) achieved by \TS and \WTS under known variance, however, as discussed in~\cite{muller17}, the expected value of the regrets asymptotically matches the lower bound predicted by~\cite{lairob85} despite the value of $\lambda>0$.
For completeness, we summarize \WTS under known $\bsigma$ in Algorithm~\ref{alg:wts_known}.
We only specify how \WTS is implemented since \TS can be directly recovered by choosing $M=1$ in Algorithm~\ref{alg:wts_known}.

%% file: figures/freq_resp_sec_WTS.pdf_tex
\begingroup%
  \makeatletter%
  \providecommand\color[2][]{%
    \errmessage{(Inkscape) Color is used for the text in Inkscape, but the package 'color.sty' is not loaded}%
    \renewcommand\color[2][]{}%
  }%
  \providecommand\transparent[1]{%
    \errmessage{(Inkscape) Transparency is used (non-zero) for the text in Inkscape, but the package 'transparent.sty' is not loaded}%
    \renewcommand\transparent[1]{}%
  }%
  \providecommand\rotatebox[2]{#2}%
  \newcommand*\fsize{\dimexpr\f@size pt\relax}%
  \newcommand*\lineheight[1]{\fontsize{\fsize}{#1\fsize}\selectfont}%
  \ifx\svgwidth\undefined%
    \setlength{\unitlength}{200bp}%
    \ifx\svgscale\undefined%
      \relax%
    \else%
      \setlength{\unitlength}{\unitlength * \real{\svgscale}}%
    \fi%
  \else%
    \setlength{\unitlength}{\svgwidth}%
  \fi%
  \global\let\svgwidth\undefined%
  \global\let\svgscale\undefined%
  \makeatother%
  \begin{picture}(1,0.55)%
    \lineheight{1}%
    \setlength\tabcolsep{0pt}%
    \put(0,0){\includegraphics[width=\unitlength,page=1]{freq_resp_sec_WTS.pdf}}%
    \put(0.38704471,0.52380109){\color[rgb]{0,0,0}\makebox(0,0)[lt]{\lineheight{1.25}\smash{\begin{tabular}[t]{l}\textbf{Means and variances}\end{tabular}}}}%
    \put(0,0){\includegraphics[width=\unitlength,page=2]{freq_resp_sec_WTS.pdf}}%
    \put(0.04709911,0.11465713){\color[rgb]{0.14901961,0.14901961,0.14901961}\makebox(0,0)[lt]{\lineheight{1.25}\smash{\begin{tabular}[t]{l}0\end{tabular}}}}%
    \put(0.49468925,0.11465713){\color[rgb]{0.14901961,0.14901961,0.14901961}\makebox(0,0)[lt]{\lineheight{1.25}\smash{\begin{tabular}[t]{l}$\pi/2$\end{tabular}}}}%
    \put(0.94868662,0.11465713){\color[rgb]{0.14901961,0.14901961,0.14901961}\makebox(0,0)[lt]{\lineheight{1.25}\smash{\begin{tabular}[t]{l}$\pi$\end{tabular}}}}%
    \put(0,0){\includegraphics[width=\unitlength,page=3]{freq_resp_sec_WTS.pdf}}%
    \put(0.03147769,0.13760111){\color[rgb]{0.14901961,0.14901961,0.14901961}\makebox(0,0)[lt]{\lineheight{1.25}\smash{\begin{tabular}[t]{l}0\end{tabular}}}}%
    \put(0.01317135,0.22867518){\color[rgb]{0.14901961,0.14901961,0.14901961}\makebox(0,0)[lt]{\lineheight{1.25}\smash{\begin{tabular}[t]{l}0.2\end{tabular}}}}%
    \put(0.01317135,0.31974925){\color[rgb]{0.14901961,0.14901961,0.14901961}\makebox(0,0)[lt]{\lineheight{1.25}\smash{\begin{tabular}[t]{l}0.4\end{tabular}}}}%
    \put(0.01317135,0.41082332){\color[rgb]{0.14901961,0.14901961,0.14901961}\makebox(0,0)[lt]{\lineheight{1.25}\smash{\begin{tabular}[t]{l}0.6\end{tabular}}}}%
    \put(0.01317135,0.50189738){\color[rgb]{0.14901961,0.14901961,0.14901961}\makebox(0,0)[lt]{\lineheight{1.25}\smash{\begin{tabular}[t]{l}0.8\end{tabular}}}}%
    \put(0,0){\includegraphics[width=\unitlength,page=4]{freq_resp_sec_WTS.pdf}}%
    \put(0.16480172,0.01712511){\color[rgb]{0.14901961,0.14901961,0.14901961}\makebox(0,0)[lt]{\lineheight{1.25}\smash{\begin{tabular}[t]{l}$|G(\ejw)|$\end{tabular}}}}%
    \put(0.46887196,0.01616896){\color[rgb]{0.14901961,0.14901961,0.14901961}\makebox(0,0)[lt]{\lineheight{1.25}\smash{\begin{tabular}[t]{l}$|H(\ejw)|$\end{tabular}}}}%
    \put(0.74672994,0.01712518){\color[rgb]{0.14901961,0.14901961,0.14901961}\makebox(0,0)[lt]{\lineheight{1.25}\smash{\begin{tabular}[t]{l}$\bmu_k = \begin{bmatrix}\Re G(\ejwk)\\\Im G(\ejwk)\end{bmatrix}$\end{tabular}}}}%
    \put(0.48558083,0.06869707){\color[rgb]{0,0,0}\makebox(0,0)[lt]{\lineheight{1.25}\smash{\begin{tabular}[t]{l}$\w$ rad/s\end{tabular}}}}%
  \end{picture}%
\endgroup%

%% file: main.bbl
\begin{thebibliography}{10}
\providecommand{\url}[1]{#1}
\csname url@samestyle\endcsname
\providecommand{\newblock}{\relax}
\providecommand{\bibinfo}[2]{#2}
\providecommand{\BIBentrySTDinterwordspacing}{\spaceskip=0pt\relax}
\providecommand{\BIBentryALTinterwordstretchfactor}{4}
\providecommand{\BIBentryALTinterwordspacing}{\spaceskip=\fontdimen2\font plus
\BIBentryALTinterwordstretchfactor\fontdimen3\font minus
  \fontdimen4\font\relax}
\providecommand{\BIBforeignlanguage}[2]{{%
\expandafter\ifx\csname l@#1\endcsname\relax
\typeout{** WARNING: IEEEtran.bst: No hyphenation pattern has been}%
\typeout{** loaded for the language `#1'. Using the pattern for}%
\typeout{** the default language instead.}%
\else
\language=\csname l@#1\endcsname
\fi
#2}}
\providecommand{\BIBdecl}{\relax}
\BIBdecl

\bibitem{thomps33}
W.~R. Thompson, ``On the likelihood that one unknown probability exceeds
  another in view of the evidence of two samples,'' \emph{Biometrika}, vol.~25,
  no.~3, pp. 285--294, 1933.

\bibitem{robbin52}
H.~Robbins, ``Some aspects of the sequential design of experiments,''
  \emph{Bulletin of the American Mathematical Society}, vol.~58, no.~5, pp.
  527--535, 1952.

\bibitem{lairob85}
T.~L. Lai and H.~Robbins, ``Asymptotically efficient adaptive allocation
  rules,'' \emph{Advances in applied mathematics}, vol.~6, pp. 4--22, 1985.

\bibitem{shen--15}
W.~Shen, J.~Wang, Y.~{G}ang Jiang, and H.~Zha, ``Portfolio choices with
  orthogonal bandit learning,'' in \emph{Twenty-Fourth International Joint
  Conference on Artificial Intelligence (IJCAI 2015)}, 2015, pp. 974--980.

\bibitem{huofu-17}
X.~Huo and F.~Fu, ``Risk-aware multi-armed bandit problem with application to
  portfolio selection,'' \emph{Royal society open science}, no. 171377, 2017.

\bibitem{talebi18}
M.~S. Talebi, Z.~Zou, R.~Combes, A.~Proutiere, and M.~Johansson, ``Stochastic
  online shortest path routing: The value of feedback,'' \emph{IEEE
  Transactions on Automatic Control}, vol.~63, no.~4, pp. 915--930, 2018.

\bibitem{kolobo20}
A.~Kolobov, S.~Bubeck, and J.~Zimmert, ``Online learning for active cache
  synchronization,'' in \emph{37th International Conference on Machine Learning
  (ICML)}, 2020, pp. 974--980.

\bibitem{press-09}
W.~H. Press, ``Bandit solutions provide unified ethical models for randomized
  clinical trials and comparative effectiveness research,'' \emph{Proceedings
  of the National Academy of Sciences}, vol. 106, no.~52, pp. 22\,387--22\,392,
  2009.

\bibitem{verstr20}
T.~Verstraeten, E.~Bargiacchi, P.~J.~K. Libin, J.~Helsen, D.~M. Roijers, and
  A.~Now\'e, ``Multi-agent {T}hompson {S}ampling for bandit applications with
  sparse neighbourhood structures,'' \emph{Nature, Scientific Reports}, p.
  6728, 2020.

\bibitem{muller17}
M.~I. M\"uller, P.~E. Valenzuela, A.~Proutiere, and C.~R. Rojas, ``A stochastic
  multi-armed bandit approach to nonparametric {$\Hinf$-norm} estimation,'' in
  \emph{Proceedings of the 56th IEEE Conference on Decision and Control (CDC)},
  2017.

\bibitem{mulroj19}
M.~I. M\"uller and C.~R. Rojas, ``Gain estimation of dynamical linear systems
  using {T}hompson {S}ampling,'' in \emph{Proceedings of the 22nd International
  Conference on Artificial Intelligence and Statistics (AISTATS)}, 2019.

\bibitem{burkat96}
A.~N. Burnetas and M.~N. Katehakis, ``Optimal adaptive policies for sequential
  allocation problems,'' \emph{Advances in Applied Mathematics}, vol.~17,
  no.~7, pp. 122--142, 1996.

\bibitem{latsze20}
T.~Lattimore and C.~Szepesvári, \emph{Bandit Algorithms}.\hskip 1em plus 0.5em
  minus 0.4em\relax Cambridge University Press, 2020.

\bibitem{auer--02}
P.~Auer, N.~Cesa-Bianchi, Y.~Freund, and R.~E. Schapire, ``The nonstochastic
  multiarmed bandit problem,'' \emph{SIAM J. Comput.}, vol.~32, no.~1, pp.
  48--77, 2002.

\bibitem{cappe-13}
O.~Cappé, A.~Garivier, O.-A. Maillard, R.~Munos, and G.~Stoltz,
  ``{K}ullback-{L}eibler upper confifence bounds for optimal sequential
  allocation,'' \emph{The Annals of Statistics}, vol.~41, no.~3, pp.
  1516--1541, 2013.

\bibitem{aumusz09}
J.-Y. Audibert, R.~Munos, and C.~Szepesv\'{a}ri, ``Exploration-exploitation
  tradeoff using variance estimates in multi-armed bandits,'' \emph{Theoretical
  Computer Science}, vol. 410, no.~19, pp. 1876--1902, 2009.

\bibitem{cowan-18}
W.~Cowan, J.~Honda, and M.~N. Katehakis, ``Normal bandits of unknown means and
  variances,'' \emph{Journal of Machine Learning Research}, vol.~18, no. 154,
  pp. 1--28, 2018.

\bibitem{chali-11}
O.~Chapelle and L.~Li, ``An empirical evaluation of {T}hompson {S}ampling,'' in
  \emph{Advances in Neural Information Processing Systems (NIPS) 24}, 2011.

\bibitem{agrgoy12}
S.~Agrawal and N.~Goyal, ``Analysis of thompson sampling for the multi-armed
  bandit problem,'' in \emph{Proceedings of the 25th Annual Conference on
  Learning Theory (COLT)}, 2012, pp. 39.1--39.26.

\bibitem{kaufma12}
E.~Kaufmann, N.~Korda, and R.~Munos, ``{T}hompson {S}ampling: An asymptotically
  optimal finite-time analysis,'' in \emph{Proceedings of the 23rd
  International Conference on Algorithmic Learning Theory (ALT)}.\hskip 1em
  plus 0.5em minus 0.4em\relax Springer, 2012, pp. 199--213.

\bibitem{korda-13}
N.~Korda, E.~Kaufmann, and R.~Munos, ``{T}hompson {S}ampling for 1-dimensional
  exponential family bandits,'' in \emph{Advances in Neural Information
  Processing Systems (NIPS) 26}, 2013.

\bibitem{hontak14}
J.~Honda and A.~Takemura, ``Optimality of {T}hompson {S}ampling for {G}aussian
  bandits depends on priors,'' in \emph{Proceedings of the 17th international
  conference on Artificial Intelligence and Statistics (AISTATS)}, 2014.

\bibitem{agrgoy12b}
S.~Agrawal and N.~Goyal, ``Further optimal regret bounds for thompson
  sampling,'' \emph{Journal of the ACM}, vol.~64, 09 2012.

\bibitem{zhotom18}
D.~P. Zhou and C.~J. Tomlin, ``Budget-constrained multi-armed bandits with
  multiple plays,'' in \emph{Proceedings of The Thirty-Second Association for
  the Advancement of Artificial Intelligence Conference (AAAI)}, 2018, pp.
  4572--4579.

\bibitem{tran-t10}
L.~Tran-Thanh, A.~Chapman, E.~Munoz~de Cote, A.~Rogers, and N.~Jennings,
  ``Epsilon–first policies for budget–limited multi-armed bandits,'' in
  \emph{Proceedings of The Twenty-Fourth Association for the Advancement of
  Artificial Intelligence Conference (AAAI)}, 2010, pp. 1211--1216.

\bibitem{wabahj10}
B.~Wahlberg, M.~B. Syberg, and H.~Hjalmarsson, ``Non-parametric methods for
  $\mathcal{L}_2$-gain estimation using iterative experiments,''
  \emph{Automatica}, vol.~46, no.~8, pp. 1376 -- 1381, 2010.

\bibitem{rojas-12}
C.~R. Rojas, T.~Oomen, H.~Hjalmarsson, and B.~Wahlberg, ``Analyzing iterations
  in identification with application to nonparametric
  {$\mathcal{H}_{\infty}$-norm} estimation,'' \emph{Automatica}, vol.~48,
  no.~11, pp. 2776--2790, 2012.

\bibitem{tubore18}
S.~Tu, R.~Boczar, and B.~Recht, ``On the approximation of {T}oeplitz operators
  for nonparametric {$\Hinf$}-norm estimation,'' in \emph{Proceedings of the
  American Control Conference (ACC)}, 2018.

\bibitem{dani--08}
V.~Dani, T.~P. Hayes, and S.~M. Kakade, ``The price of bandit information for
  online optimization,'' in \emph{Proceedings of the 21st Annual Conference on
  Learning Theory (COLT)}, 2008, pp. 355--366.

\bibitem{bubces12}
S.~Bubeck and N.~Cesa-Bianchi, ``Regret analysis of stochastic and
  nonstochastic multi-armed bandit problems,'' \emph{Foundations and Trends in
  Machine Learning}, vol.~5, no.~1, pp. 1--122, 2012.

\bibitem{kaufma14}
E.~Kaufmann, ``Analyse de strat\'egies bay\'esiennes et fr\'equentistes pour
  l'allocation s\'equentielle de ressources,'' Ph.D. dissertation, Paris
  Institute of Technology, Paris, France, 2014.

\bibitem{papoul91}
A.~Papoulis and U.~Pillai, \emph{Probability, Random Processes, and Stochastic
  Processes}, 4th~ed.\hskip 1em plus 0.5em minus 0.4em\relax McGraw-Hill, 2002.

\bibitem{zhodoy96}
K.~Zhou, J.~C. Doyle, and K.~Glover, \emph{Robust and Optimal Control}.\hskip
  1em plus 0.5em minus 0.4em\relax Prentice-Hall, 1996.

\bibitem{ljung-99}
L.~Ljung, \emph{System Identification: Theory for the User}, 2nd~ed.\hskip 1em
  plus 0.5em minus 0.4em\relax Prentice Hall, 1999.

\bibitem{vdscha17}
A.~van~der Schaft, \emph{$L_2$ Gain and Passivity Techniques in Nonlinear
  Control}, 3rd~ed.\hskip 1em plus 0.5em minus 0.4em\relax Springer, 2017.

\bibitem{muller18}
M.~I. Müller, J.~Milošević, H.~Sandberg, and C.~R. Rojas, ``A
  risk-theoretical approach to $\mathcal{H}_{2}$-optimal control under covert
  attacks,'' in \emph{Proceedings of the 57th IEEE Conference on Decision and
  Control (CDC)}, 2018, pp. 4553--4558.

\bibitem{kay---93}
S.~M. Kay, \emph{Fundamentals of Statistical Signal Processing: Estimation
  Theory}.\hskip 1em plus 0.5em minus 0.4em\relax Prentice-Hall, 1993.

\bibitem{fairma98}
F.~W. Fairman, \emph{Linear Control Theory: The State Space Approach}.\hskip
  1em plus 0.5em minus 0.4em\relax John Willey \& Sons, 1998.

\bibitem{kailat80}
T.~Kailath, \emph{Linear Systems}.\hskip 1em plus 0.5em minus 0.4em\relax
  Prentice Hall, 1980.

\bibitem{oppsch99}
A.~V. Oppenheim and R.~W. Schafer, \emph{Discrete-time Signal
  Processing}.\hskip 1em plus 0.5em minus 0.4em\relax Prentice-Hall, 2009.

\bibitem{aguero10}
J.~C. Ag{\"u}ero, J.~I. Yuz, G.~C. Goodwin, and R.~Delgado, ``On the
  equivalence of time and frequency domain maximum likelihood estimation,''
  \emph{Automatica}, vol.~46, no.~2, pp. 260--270, 2010.

\bibitem{covtho06}
T.~M. Cover and J.~A. Thomas, \emph{Elements of Information Theory},
  2nd~ed.\hskip 1em plus 0.5em minus 0.4em\relax Wiley-Interscience, 2006.

\bibitem{hardy-88}
G.~H. Hardy, J.~E. Littlewood, and G.~P\'olya, \emph{Inequalities},
  2nd~ed.\hskip 1em plus 0.5em minus 0.4em\relax Cambridge University Press,
  1988.

\bibitem{durret10}
R.~Durrett, \emph{Probability: Theory and Examples}, 4th~ed.\hskip 1em plus
  0.5em minus 0.4em\relax Cambridge University Press, 2010.

\bibitem{horjoh12}
R.~A. Horn and C.~R. Johnson, \emph{Matrix Analysis}, 2nd~ed.\hskip 1em plus
  0.5em minus 0.4em\relax Cambridge University Press, 2012.

\bibitem{soders02}
T.~S\"oderstr\"om, \emph{Discrete-time Stochastic Systems}.\hskip 1em plus
  0.5em minus 0.4em\relax Springer, 2002.

\bibitem{fishma96}
G.~Fishman, \emph{Monte {C}arlo: Concepts, Algorithms, and Applications}.\hskip
  1em plus 0.5em minus 0.4em\relax Springer, 1996.

\end{thebibliography}
